\documentclass[10pt,twocolumn,letterpaper]{article}

\usepackage[pagenumbers]{cvpr} %

\definecolor{cvprblue}{rgb}{0.21,0.49,0.74}
\usepackage[pagebackref,breaklinks,colorlinks,allcolors=cvprblue]{hyperref}

\usepackage{orcidlink}

\usepackage{url}            %
\usepackage{amsfonts}       %
\usepackage{nicefrac}       %
\usepackage{microtype}      %

\usepackage{amsmath,amssymb, bm} %
\usepackage{epsfig}
\usepackage{multirow}
\usepackage{graphicx}
\usepackage{mathtools}
\usepackage{wrapfig}
\usepackage{bbm}
\usepackage{gensymb}
\usepackage{comment}
\usepackage{algpseudocode}
\usepackage{color}
\usepackage{here}
\usepackage{makecell}
\usepackage{adjustbox} %
\usepackage{array}
\usepackage{multirow}
\usepackage{rotating}
\usepackage{float}
\usepackage{caption} 
\usepackage{xr}
\def\paperID{12324} %
\def\confName{CVPR}
\def\confYear{2025}

\title{GA3CE: Unconstrained 3D Gaze Estimation with \\ Gaze-Aware 3D Context Encoding}

\author{
Yuki Kawana \quad 
Shintaro Shiba \quad 
Quan Kong \quad 
Norimasa Kobori \\
Woven by Toyota\\
}

\begin{document}
\twocolumn[{
\renewcommand\twocolumn[1][]{#1}
\maketitle
\begin{center}
    \includegraphics[width=\linewidth]{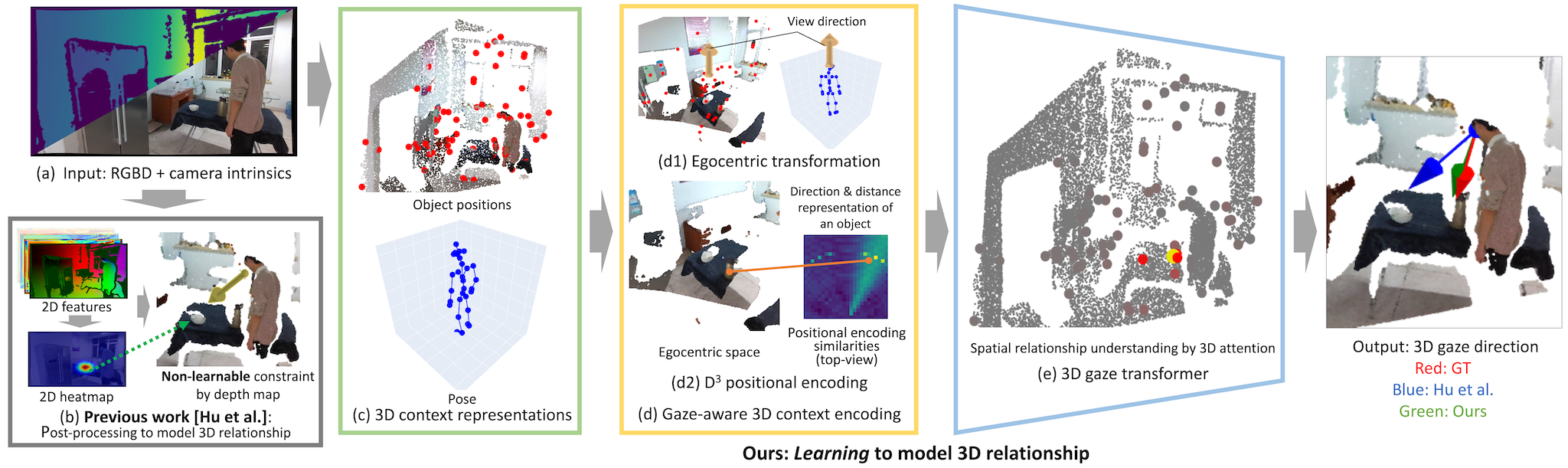}
   \vspace{-1.5\baselineskip}
    \captionof{figure}{
(a) Our method estimates 3D gaze direction from an RGBD image and camera intrinsics.  
(b) Prior work \cite{hu2023gfie} estimates gaze from 2D representations, incorporating 3D spatial cues only as a non-learnable post-processing step.  
(c) Since direct 3D gaze estimation from 2D representations is challenging, we use 3D pose and object positions as intermediate representations.  
(d) We introduce gaze-aware 3D context encoding (GA3CE), transforming the 3D context into a unified directional space.  
(d1) This space aligns with a egocentric (subject-centered) view, normalized relative to the head position and oriented to the view direction.  
(d2) This alignment enables the decomposition of a 3D point into direction and distance in egocentric space, with D$^3$ positional encoding capturing their correlations.
(e) The transformer then learns spatial relationships between the subject, objects, and 3D gaze.
    }
    \label{fig:intro}
\vspace{0.5\baselineskip}

\end{center}
}]
\maketitle

\begin{abstract}
We propose a novel 3D gaze estimation approach that learns spatial relationships between the subject and objects in the scene, and outputs 3D gaze direction. Our method targets unconstrained settings, including cases where close-up views of the subject's eyes are unavailable, such as when the subject is distant or facing away. Previous approaches typically rely on either 2D appearance alone or incorporate limited spatial cues using depth maps in the non-learnable post-processing step. Estimating 3D gaze direction from 2D observations in these scenarios is challenging; variations in subject pose, scene layout, and gaze direction, combined with differing camera poses, yield diverse 2D appearances and 3D gaze directions even when targeting the same 3D scene. To address this issue, we propose GA3CE: Gaze-Aware 3D Context Encoding. Our method represents subject and scene using 3D poses and object positions, treating them as 3D context to learn spatial relationships in 3D space. Inspired by human vision, we align this context in an egocentric space, significantly reducing spatial complexity. Furthermore, we propose D$^3$ (direction-distance-decomposed) positional encoding to better capture the spatial relationship between 3D context and gaze direction in direction and distance space. Experiments demonstrate substantial improvements, reducing mean angle error by 13\%–37\% compared to leading baselines on benchmark datasets in single-frame settings. \noindent Project page: \href{https://woven-visionai.github.io/ga3ce-project/}{\texttt{woven-visionai.github.io/ga3ce-project}}
\end{abstract}

\section{Introduction}
\label{sec:intro}
Gaze direction is a powerful non-verbal cue for understanding human engagement, interest, and attention. Humans can often infer another person's gaze direction based on various appearance cues, even when clear, close-up views of the eyes are not available, such as when the person is distant or facing away. This is possible because humans interpret gaze through cues like body pose and movement, scene context, and spatial relationships. Estimating 3D gaze direction in such unconstrained settings has a range of applications. Consider scenarios where wearable eye-tracking devices are impractical, such as detecting whether a pedestrian is attentive to traffic via a surveillance camera \cite{kong2024wts,benfold2011unsupervised} or analyzing customer engagement in a retail environment using video monitoring \cite{wang2022gatector,wang2024transgop}.

To estimate 3D gaze direction in these unconstrained settings, previous work has utilized cues that imply gaze direction, such as a subject's temporal 3D direction \cite{nonaka2022dynamic}. Recently, depth maps have been incorporated to analytically constrain 3D gaze direction estimated from head appearance \cite{hu2023gfie}. 
However, previous works overlook spatial relationships between the subject, scene, and gaze direction; scene context is not considered \cite{nonaka2022dynamic} or only considered during non-learnable, analytical step \cite{hu2023gfie}. Therefore, an effective learning approach that holistically considers both subject and scene for 3D gaze estimation remains largely unexplored.

In this work, we propose a novel 3D gaze estimation approach that learns to understand the spatial relationship between the scene, subject, and gaze.
Following the closest previous work \cite{hu2023gfie}, our method estimates 3D gaze direction given an RGB image, a depth map of the scene from a sensor or a zero-shot estimator, and camera intrinsics. In contrast to the gaze-following task which focuses on detecting the visible gazed \emph{point}, our focus is to estimate the 3D \emph{direction} of gaze. The task has orthogonal benefit to the gaze-following task, as it can output gaze information even when an attended point is not visible due to occlusion or being out of sight. 

Estimating 3D gaze direction from 2D observations is challenging due to variations in 2D scene context and gaze direction. Even with identical 3D scenes, 2D appearances of the subject and scene, and ground truth gaze direction defined in camera space, change significantly with different camera poses, as each pose projects the 3D world differently onto a 2D image. This creates complex interactions between the relative positions of 2D features and 3D gaze directions across varying camera poses. Normalizing close-up 2D facial images with perspective correction to reduce appearance variation from different camera poses has been widely studied for 3D gaze estimation \cite{sugano2014learning,zhang2018revisiting}. However, the normalization beyond the subject's 2D face appearance remains largely unexplored.

To address the complexity of 3D gaze estimation, we focus on three key challenges in this paper:
($i$) What is an effective representation of the subject and scene?
($ii$) How can variations in subject, scene, and gaze due to different camera poses be normalized?
($iii$) How can we model the spatial relationships among subject, scene, and gaze?

For ($i$), we represent the subject as 3D keypoints and the scene as 3D points corresponding to object positions. Instead of estimating 3D gaze direction directly from the 2D appearance of the subject and scene, these serve as robust intermediate 3D context representations. For ($ii$) and ($iii$), inspired by human vision studies \cite{ma2015learning, einhauser2008objects, findlay2003active} demonstrating that direction and distance to objects in the scene within subject's view strongly influences gaze, we propose \emph{GA3CE}, for Gaze-Aware 3D Context Encoding. To address ($ii$), this intermediate 3D representation enables geometric transformation of 3D contexts relative to the subject's view direction, such as prior gaze or head direction. This transformation as normalization reduces context variations across camera poses, creating a unified, egocentric space. To address ($iii$), we decompose the 3D context into direction and distance components for positional encoding, termed D$^3$ positional encoding, which better captures positional and directional similarities. This decomposition, combined with a transformer-based architecture, models the spatial interactions between 3D context and the final 3D gaze direction. An overview of our method is shown in \cref{fig:intro}.

Experiments on three benchmark datasets demonstrate that our approach significantly outperforms previous methods, and an ablation study highlights the advantages of using a 3D representation with GA3CE for learning spatial relationships. Our contributions are summarized as follows:
\begin{enumerate}
\item We propose a novel approach for 3D gaze estimation based on an explicit understanding of the spatial relationships between a subject and objects in a 3D context.
\item We propose \emph{GA3CE}, for Gaze-Aware 3D Context Encoding, to enhance the representation of the relationship between 3D context and gaze.
\item We achieve state-of-the-art quantitative and qualitative performance on three benchmark datasets \cite{hu2023gfie, nonaka2022dynamic, Koppula2013}.
\end{enumerate}

\begin{figure*}[t]
    \begin{center}
        \includegraphics[width=\linewidth]{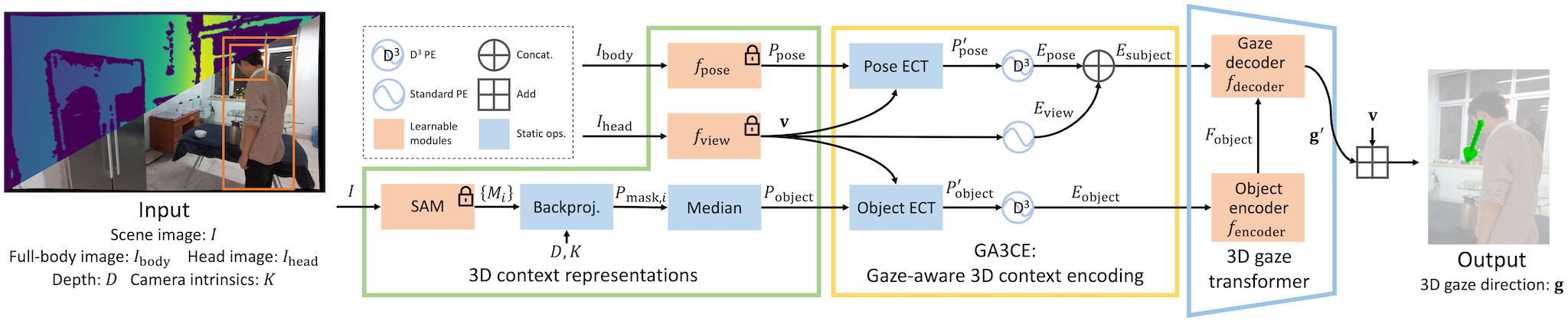}
    \end{center}
   \vspace{-1.5\baselineskip}
    \caption{Pipeline overview. PE = positional encoding; ECT = egocentric transformation. First, we extract 3D pose and object positions as 3D context representations. To reduce their variation and better capture spatial relationships between 3D context and gaze direction, we apply GA3CE: gaze-aware 3D context encoding. Using egocentric transformation, we convert 3D context to a egocentric space, and encoding them into a high-dimensional feature space for direction and distance with D$^3$ positional encoding. Finally, the 3D gaze transformer learns spatial relationships between the 3D context and gaze.}
    \label{fig:pipeline-overview}
\vspace{-0.5\baselineskip}
\end{figure*}

\section{Related Work}
\paragraph{3D gaze estimation.}
Geometry-based 3D gaze estimation approaches have long been studied in computer vision. They use a 3D eye model to regress the 3D gaze direction based on geometric and optical characteristics of the eye's appearance \cite{hennessey2006single,lee20123d}, achieving high accuracy in direction estimation at the cost of requiring eye trackers.

Learning-based approaches that estimate the 3D gaze direction from a subject's appearance have also been explored, from a close-up view of the face \cite{fischer2018rt,zhang2020eth,zhang2015appearance,zhang2017s} or under an unconstrained setting where the face is not clearly visible because the subject is facing backward \cite{kellnhofer2019gaze360}. More recent works \cite{nonaka2022dynamic,hu2022we,hu2023gfie} target 3D gaze estimation in challenging scenarios where the subject is distant from the camera and captured under various camera poses.

In this paper, we also target the challenging scenario where we do not have a clear view of the subject's facial features, and images are captured under diverse camera poses w.r.t. the subject, resulting in diverse subject's appearances and 3D gaze directions.

\paragraph{Gaze estimation by context cues.}
Estimating 3D gaze based on context cues has been explored using optimization-based methods with hand-crafted energy terms on temporal RGB(D) data, assuming known object positions and categories as context cues \cite{wei2017inferring, wei2018and, brau2018multiple}.

In learning-based approaches, previous works on 2D gaze-following tasks, which localize the gazed point in an image, have proposed using various context cues, including body pose \cite{guan2020enhanced, gupta2022modular}, depth \cite{fang2021dual, tafasca2023childplay, gupta2022modular}, objects \cite{tonini2023object, tafasca2024toward, yang2024gaze}, actions \cite{yang2024gaze}, and speech \cite{hou2024multi}. Some of these studies \cite{tonini2023object, tafasca2023childplay, gupta2022modular} estimate the 3D gaze direction solely from head appearance as one of the context cues, while the other cues are used only for the final 2D gaze-following.

In 3D gaze estimation, prior work \cite{nonaka2022dynamic} utilizes motion cues by leveraging temporal appearances and 2D body flow. Another study \cite{hu2022we} uses depth maps as 2D features to estimate 3D gaze direction as a prior for 3D gaze-following. The latest and the closest work to ours \cite{hu2023gfie} estimates 3D gaze direction in a post-processing step using depth maps for directional constraints and gaze-following modules for 2D scene understanding.

In contrast, our approach directly learns to estimate 3D gaze direction by modeling the spatial relationships of 3D context, without relying on 2D gaze-following or post-processing for 3D understanding. Furthermore, this work proposes a unified approach that integrates 3D pose and object-level scene understanding for 3D gaze estimation, while the use of context cues has been limited to 2D gaze understanding in previous works.

\section{Method}
Given an RGB image $I$ with the subject of interest, the corresponding depth map $D$ from either a depth sensor or zero-shot depth estimator, and known camera intrinsics $K$ for each image, our method estimates the subject's 3D gaze direction $\mathbf{g}\in\mathbb{S}^2$. We assume 2D bounding boxes for the head $\mathbf{b}_{\text{head}}$ and the full-body $\mathbf{b}_{\text{body}}$ of the subject are provided, as in the previous works \cite{hu2023gfie,nonaka2022dynamic}.

Our approach is outlined in \cref{fig:pipeline-overview}. First, we extract 3D human pose and object positions as 3D context representations (\cref{sec:contextRepr}). Next, GA3CE strengthens the learning of spatial relationships between gaze direction and 3D context (\cref{sec:sub-cent}). Finally, the 3D gaze transformer estimates gaze direction by modeling spatial relationships within the input context (\cref{sec:transformer}).
\subsection{3D context representations}
\label{sec:contextRepr}

\paragraph{Subject representation.}
We represent the subject using 3D body keypoints as a pose, and a view direction as a 3D unit vector. Both body direction \cite{nonaka2022dynamic} and pose \cite{wei2017inferring,wei2018and} are closely related to gaze direction. Building on these insights, we incorporate 3D human pose estimation. Specifically, we use the pre-trained 3D human pose estimator \cite{Sarandi2023dozens}, \( f_{\text{pose}} \), which takes a single RGB image, \( I_{\text{body}} \), cropped from \( I \) using \( \mathbf{b}_{\text{body}} \), as input to estimate 3D keypoints. The output, \( P_{\text{pose}} = f_{\text{pose}}(I_{\text{body}}) \in \mathbb{R}^{N_{\text{pose}} \times 3} \), represents the keypoints in camera space, where \( N_{\text{pose}} \) is the number of keypoints. Since the estimator's output scale and translation may not align precisely with the input depth map \( D \), a depth-aware human pose estimator \cite{zimmermann20183d} or learning-based post-processing \cite{bashirov2021real} could be considered. However, these approaches are costly to train across varying depth maps or require resource-intensive refinement steps. Instead, we use the human pose estimator trained on large-scale RGB datasets, which generalizes effectively across domains. In practice, we observed the proposed pipeline adapting to different scales and translations between 3D poses and depth maps without issue. Following the previous work \cite{hu2023gfie}, we use the pre-trained appearance-based estimator \( f_\text{view} \) to estimate the view direction \(\mathbf{v} = f_\text{view}(I_\text{head}) \in \mathbb{S}^2\) as a directional prior. The input is the subject's head image \( I_\text{head} \), cropped from \( I \) using \( \mathbf{b}_{\text{head}} \).

\begin{figure}[t]
    \begin{center}
    \includegraphics[width=\columnwidth]{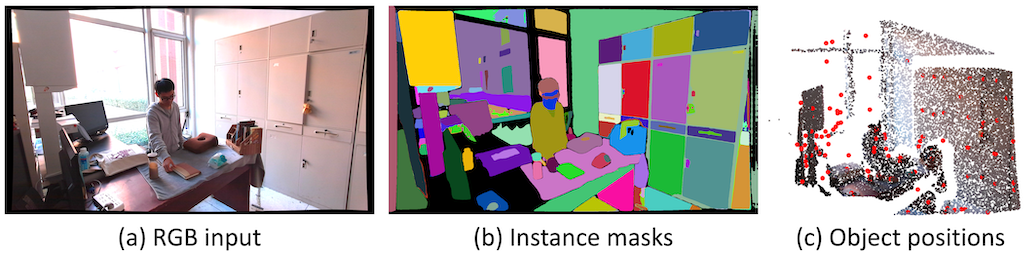}
    \end{center}
   \vspace{-1.5\baselineskip}
    \caption{Illustration of object position sampling using the segment-everything approach \cite{mobile_sam}. This method comprehensively identifies object positions in the scene, shown as red points in (c). The colored point cloud is shown for visualization purposes only.
}
    \label{fig:sam-vis}
\vspace{-0.25\baselineskip}
\end{figure}

\paragraph{Object representation.} 
The presence of objects in a scene can substantially influence the direction of a subject's gaze \cite{tonini2023object,wei2017inferring,wei2018and,brau2018multiple}. Previous studies, however, either assume known 3D object positions or require 2D instance annotations for training, and are often limited to specific categories. To address these limitations, we use the Segment Anything (SAM) framework \cite{Kirillov_2023_ICCV} to sample object positions in the scene, visualized in \cref{fig:sam-vis}. 

We represent object positions \( P_{\text{object}} \in \mathbb{R}^{N_{\text{object}}\times3} \) as their 3D coordinates in the scene, where \( N_\text{object} \) is the number of detected objects. For each \(i\)-th object, we first obtain a 2D instance mask \( M_i \) from SAM. We then backproject \( M_i \) into camera space as 3D points \( P_{\text{mask},i} \) using the camera intrinsics \( K \) and corresponding depth values based on the pinhole camera model. The object positions \( P_\text{object} \) are determined by taking the median of these 3D points \( P_{\text{mask}, i} \). 

The original SAM \cite{Kirillov_2023_ICCV} typically requires 2D points as prompts to sample instance masks, where mask granularity depends on the density of 2D points in the image. To comprehensively sample objects of varying sizes, the previous work \cite{qin2023langsplat} employs multi-resolution point sampling as a few-shot process. To simplify this, we use MobileSAM \cite{mobile_sam}, which provides exhaustive instance masks through a \emph{segment-everything} approach in a single shot. Note that we do not consider object appearance features in this study. Including category-agnostic semantic features, such as CLIP \cite{radford2021learning}, is left for future work.

\subsection{Gaze-aware 3D context encoding (GA3CE)}
\label{sec:sub-cent}
The subject's 3D pose, object positions, and gaze direction form a complex interaction. To effectively model this relationship, we propose a novel 3D context encoding approach tailored for understanding 3D gaze direction.

A key insight is that gaze fixations tend to focus around the center of one's view, with depth information enhancing gaze saliency prediction in first-person perspectives \cite{ma2015learning,einhauser2008objects,findlay2003active}. This indicates that the direction and distance of objects relative to the subject's view are strong cues for estimating gaze direction. Guided by this idea, we (1) normalize the 3D positions of objects $P_\text{object}$ and the 3D pose $P_\text{pose}$ within a geometrically aligned, egocentric 3D coordinate space, and (2) decompose them into direction and distance components for positional encoding.

\begin{figure}[t]
    \begin{center}
    \includegraphics[width=\columnwidth]{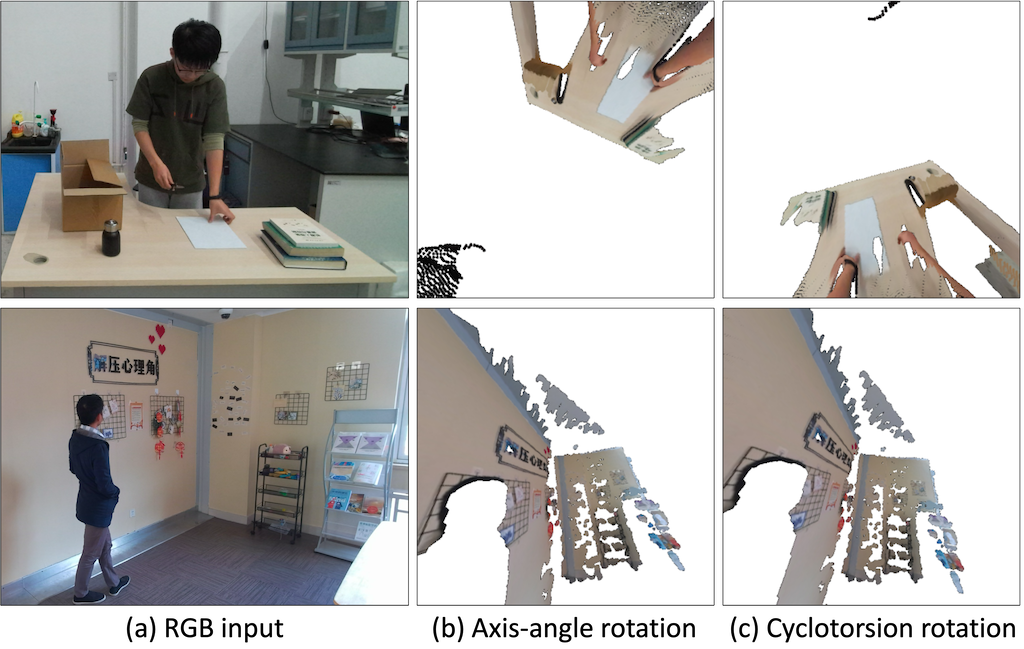}
    \end{center}
   \vspace{-1.5\baselineskip}
   \caption{
   Visualization of geometric normalization in 3D context through the egocentric transformation. (b) and (c) show 2D renderings of the colored point cloud after applying the egocentric transformation with different rotation normalizations. Note that, for intuitive visualization of the figure, we used the colored point cloud and its 2D rendering instead of the 3D pose and object positions. They are used solely for visualization purpose.
   }
    \label{fig:sct-vis}
\vspace{-0.25\baselineskip}
\end{figure}

\paragraph{Egocentric transformation.}
Unlike the previous works that focus solely on 2D features \cite{hu2023gfie,tonini2023object,nonaka2022dynamic}, our approach explicitly considers a 3D representation of the input. This enables us to align the input context to an egocentric space, simplifying the complex relationship between input contexts and the output 3D gaze direction due to varying camera poses. Without this geometric normalization, the network would need to learn these complex combinations.

We normalize the 3D pose and object positions to be relative to the head position of the subject. Specifically, we normalize the 3D pose \( P_\text{pose} \) relative to the head position \( \mathbf{t}_\text{pose} \in \mathbb{R}^3 \) and scale it using head width \( s \in \mathbb{R}^+ \).  Additional details are provided in \cref{sec:ap-pose}. Object positions \( P_\text{object} \) are also normalized relative to the camera center, adjusting them to be relative to the subject's head position \( \mathbf{t}_\text{object} \in \mathbb{R}^3 \), determined via backprojection using the center of the head bounding box $\mathbf{b}_\text{head}$, corresponding depth, and camera intrinsics \( K \). Note that both $\mathbf{t}_\text{pose}$ and $\mathbf{t}_\text{object}$ represent head position but differ in value, with one derived from \( P_\text{pose} \) and the other from the depth map \( D \).

Next, we align the rotation of the 3D pose and object positions by deriving a rotation \( R \in \text{SO}(3) \) so that the subject's view direction \( \mathbf{v} \) in camera space aligns with the fixed direction \( \mathbf{z} = R \mathbf{v} = [0, 0, 1] \). This egocentric transformation is visualized in \cref{fig:sct-vis}. The rotation can be derived as an axis-angle rotation around the axis \( \mathbf{z} \times \mathbf{v} \) by \( \arccos(\mathbf{z}^T \mathbf{v}) \). However, this method results in inconsistent rotation along the z-axis depending on \( \mathbf{v} \), as illustrated in \cref{fig:sct-vis} (b). A mathematical explanation is provided in \cref{sec:ap-rot-aa}.
Inspired by cyclotorsion, the reactive eye movement that counteracts the head tilt to keep the horizontal axis of vision aligned with the horizon, we propose a modified approach called cyclotorsion rotation. Assuming that the horizon appears horizontal in the image $I$, then the rotation $R$ is defined as:
\begin{equation}
\label{eq:in-plane}
R = \text{Euler}(\theta, \phi, 0) \in \text{SO}(3) \ \text{s.t.} \ \min_{\theta,\phi} \Vert R \mathbf{v} - \mathbf{z} \Vert.
\end{equation}
Details and the analytical solution are provided in \cref{sec:ap-rot,sec:ap-rot-cr}. Cyclotorsion rotation achieves more consistent egocentric transformation, as shown in \cref{fig:sct-vis} (c).

In summary, the egocentric transformation for view direction $\mathbf{v}$, pose $P_\text{pose}$ and object positions $P_\text{object}$ reads:
\begin{equation}
\label{eq:normalize}
\begin{split}
\mathbf{v}' &=\mathbf{z}= R\mathbf{v} \\
P_\text{pose}' &= sR(P_\text{pose} - \mathbf{t}_\text{pose}) \\
P_\text{object}' &= R(P_\text{object} - \mathbf{t}_\text{object}).
\end{split}
\end{equation}
\begin{figure}[t]
    \begin{center}
    \includegraphics[width=\columnwidth]{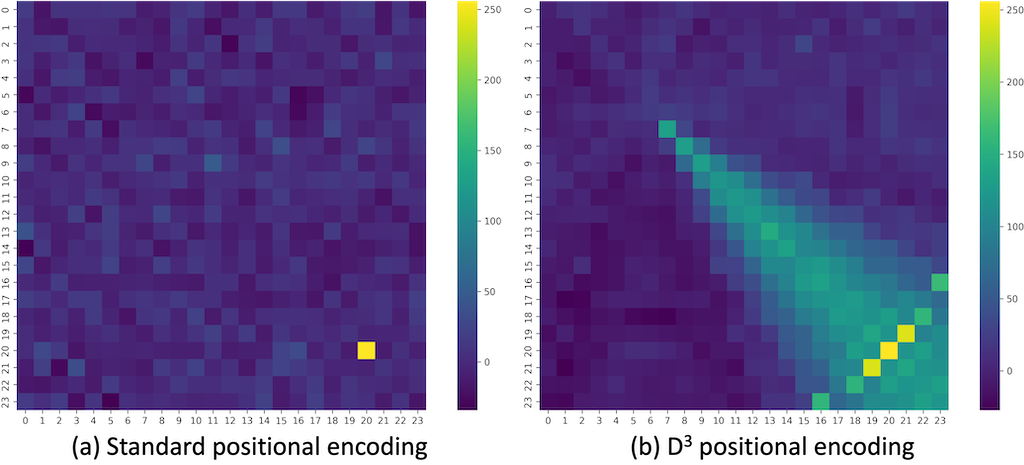}
    \end{center}
   \vspace{-1.5\baselineskip}
    \caption{2D visualization of D$^3$ positional encoding. Each point on a grid shows a dot product between the encoded reference point $\mathbf{x}_\text{ref}=(20, 20)$ and an encoded point $\mathbf{x}\in\Omega$ on a 2D grid as an unnormalized similarity score. (a) shows the standard positional encoding \cite{tancik2020fourier}, where similarity is high only at the reference point $\mathbf{x}_\text{ref}$ itself. (b) displays the proposed D$^3$ positional encoding, capturing both positional and directional similarities defined as $\tilde{\gamma}(\mathbf{x}_\text{ref}-\mathbf{c})^T\tilde{\gamma}(\mathbf{x}-\mathbf{c})$, given $\mathbf{c}=(6, 6)$ as the origin. This creates a radial gaze pattern from the origin $\mathbf{c}$ toward the reference point $\mathbf{x}_\text{ref}$, with similarity gradually increasing along the direction and distance from the origin, simulating a gaze effect.
    }
    \label{fig:d3-vis}
\end{figure}

Note that there exist exocentric-to-egocentric view synthesis methods which learns transformation from ego-exo images \cite{liu2020exocentric, liu2021cross, luo2024put} or leverages known transformations \cite{cheng20244diff}. In contrast, our egocentric transformation neither requires training nor assumes a known transformation.

\paragraph{D$^3$ positional encoding.}
The 3D direction and distance information in a subject's view significantly influence human attention \cite{ma2015learning,einhauser2008objects,findlay2003active}. Thus, we hypothesize that features explicitly capturing similarities in both direction and distance improve 3D gaze estimation. With this idea, we introduce D$^3$ (direction-distance-decomposed) positional encoding.

We employ the commonly used sinusoidal functions \cite{tancik2020fourier} as our standard positional encoding $\gamma$. We define the D$^3$ positional encoding of a 3D point $\mathbf{p}$ as $\tilde{\gamma}(\mathbf{p})= \gamma(\frac{\mathbf{p}}{\Vert \mathbf{p} \Vert}) \oplus \gamma({\Vert \mathbf{p} \Vert})$, where $\oplus$ denotes vector concatenation. We compare the standard positional encoding with the proposed D$^3$ positional encoding in \cref{fig:d3-vis}. For simplicity, we denote the standard and D$^3$ positional encodings applied to each point in the set \( P \) as \( \gamma(P) \) and \( \tilde{\gamma}(P) \), respectively. Thus, the positional encodings for view direction, pose, and objects are:
\begin{equation}
\label{eq:rpos}
\begin{split}
E_\text{view} &= \gamma_\text{view}(\mathbf{v}') \in \mathbb{R}^{C_\text{gaze}} \\
E_\text{pose} &= \tilde{\gamma}_\text{pose}(P_\text{pose}') \in \mathbb{R}^{N_\text{pose} \times C_\text{keypoint}} \\
E_\text{object} &= \tilde{\gamma}_\text{object}(P_\text{object}') \in \mathbb{R}^{N_\text{object} \times C_\text{latent}}
\end{split}
\end{equation}
where $C_*$ is the dimension of the encoded embedding. Note that although \( E_\text{view} \) remains constant when applying normalization, where \( \mathbf{v}' = \mathbf{z} = R\mathbf{v} \) as defined in \cref{eq:normalize}, we still apply positional encoding to ensure architectural consistency in the ablation study \cref{sec:abl} without normalization, where \( \mathbf{v}' = \mathbf{v} \) and \( E_\text{view} \) is not constant.
Details of $\gamma_\text{view}$, $\tilde{\gamma}_\text{pose}$, and $\tilde{\gamma}_\text{object}$ can be found in \cref{sec:supp-pe}. Note that the 3D points $P_\text{pose}'$ and $P_\text{object}'$ are relative to their respective head positions $\mathbf{t}_\text{pose}$ and $\mathbf{t}_\text{object}$, allowing D$^3$ positional encoding to represent high-dimensional direction and distance embedding relative to the head position. We define the subject embedding as \( E_\text{subject} = E_\text{view} \oplus E_\text{pose} \in \mathbb{R}^{C_\text{latent}} \), where \( E_\text{pose} \) is flattened for concatenation to form a 1D embedding of size \( C_\text{pose} = N_\text{pose} \times C_\text{keypoint} \), and \( C_\text{latent} = C_\text{gaze} + C_\text{pose} \).

\subsection{3D gaze transformer}
\label{sec:transformer}
To capture the relationship between the subject and objects, we use a transformer architecture \cite{vaswani2017attention}. Detailed descriptions of the encoder and decoder components can be found in \cite{vaswani2017attention}. The transformer encoder, \( f_\text{encoder} \), employs \( N_\text{encoder} \) layers of self-attention and non-linear transformations to encode object features \( F_\text{object} \) from the object embedding \( E_\text{object} \). The transformer decoder, \( f_\text{decoder} \), applies \( N_\text{decoder} \) layers of self-attention, non-linear transformations, and cross-attention to decode the residual 3D gaze direction \( \mathbf{g}' \) in an egocentric space.
In the cross-attention, the subject embedding \( E_\text{subject} \) is the query, and the object feature \( F_\text{object} \) serves as both key and value, producing the gaze feature as a weighted sum of relevant object features. This allows the model to focus on the relevant object positions.

Finally, the 3D gaze direction \( \mathbf{g} \) is obtained by reversing the normalization: \( \mathbf{g} = R^T \mathbf{g}' + \mathbf{v} \). We normalize $\mathbf{g}$ to have unit length. Architecture details can be found in \cref{sec:ap-transformer}.

We train the transformer network by minimizing the angular error between the predicted 3D gaze direction \( \mathbf{g} \) and the ground truth \( \mathbf{g}_\text{GT} \), defined as \( \mathcal{L} = \text{arccos}(\mathbf{g}^T \mathbf{g}_\text{GT}) \).

\begin{figure*}[t]
    \begin{center}
    \includegraphics[width=1.0\linewidth]{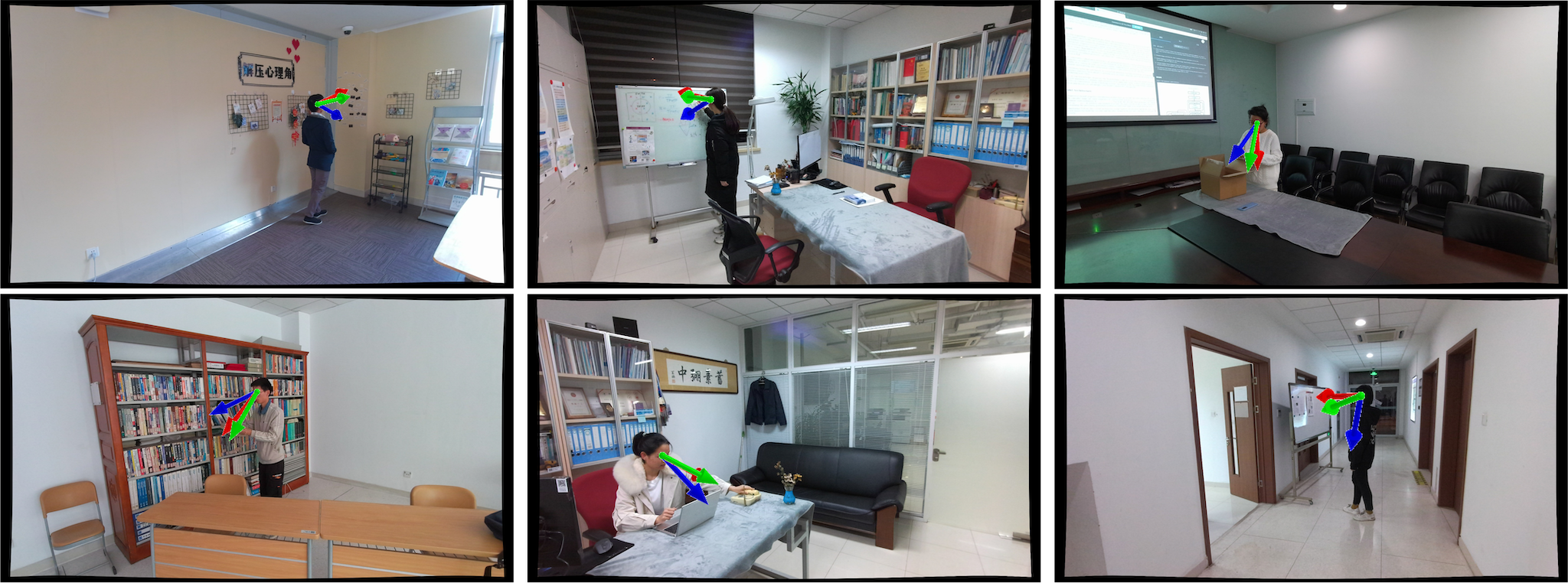}
    \end{center}
   \vspace{-1.5\baselineskip}
\caption{Qualitative results on the GFIE dataset \cite{hu2023gfie}. Red, green, and blue arrows indicate the ground truth, Ours, and GFIE \cite{hu2023gfie}.}
    \label{fig:gfie-vis}
     \vspace{-.5\baselineskip}
\end{figure*}

\begin{figure*}[t]
    \begin{center}
    \includegraphics[width=1.0\linewidth]{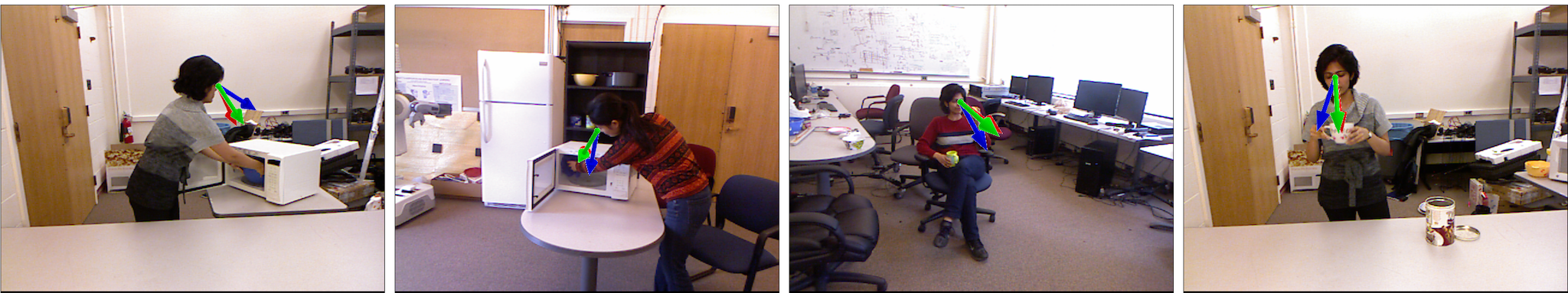}
\end{center}
\vspace{-1.5\baselineskip}
\caption{Qualitative results on the CAD-120 dataset \cite{Koppula2013}. Red, green, and blue arrows shows the ground truth, Ours + GFM, and GFIE \cite{hu2023gfie}.}
        
    \label{fig:cad120-vis}
\vspace{-1.0\baselineskip}
\end{figure*}

\subsection{Implementation details}
\label{sec:imp}
We use a pre-trained model for each dataset as the view direction estimator \( f_\text{view} \). More details on \( f_\text{view} \) are provided in \cref{sec:exp}. The weights of the SAM module \cite{mobile_sam}, \( f_\text{view} \), and the 3D human pose estimator \( f_\text{pose} \) \cite{Sarandi2023dozens} are frozen and remain unchanged during training. We set \( N_\text{pose} = 15 \), while \( N_\text{object} \) depends on the number of instances detected by MobileSAM in each input. The embedding dimensions are set as follows: \( C_\text{latent} = 256 \), \( C_\text{gaze} = 106 \), and $C_\text{keypoint}=10$. The object encoder \( f_\text{encoder} \) and gaze decoder \( f_\text{decoder} \) each consist of \( N_\text{encoder} = 3 \) transformer encoder layers and \( N_\text{decoder} = 3 \) transformer decoder layers, respectively. We train the network for 20 epochs on a single A10G GPU, using the AdamW optimizer with a learning rate of 0.0014. Further details are provided in \cref{sec:ap-training}.
\begin{table}[t]
\begin{center}
\resizebox{\columnwidth}{!}{
\begin{tabular}{lcccccc}

\hline
Method & Input & 2D gaze & AUC $\uparrow$ & L2 Dist. $\downarrow$ & 3D Dist. $\downarrow$ & 3D MAE $\downarrow$ \\
\hline
Random & & & 0.585 & 0.425 & 2.930 & 84.4 \\
Center & & & 0.614 & 0.287 & 2.510 & 87.2 \\
Rt-Gene \cite{fischer2018rt} & H & & 0.823 & 0.123 & 0.552 & 21.0 \\
Gaze360 \cite{kellnhofer2019gaze360} & H & & 0.821 & 0.130 & 0.540 & 19.8 \\
GazeFollow \cite{recasens2015they} & SD &\checkmark & 0.941 & 0.131 & 0.856 & 41.5 \\
Lian \cite{lian2018believe} & SD &\checkmark & 0.962 & 0.091 & 0.542 & 26.7 \\
Chong \cite{chong2020detecting} & SD &\checkmark & 0.972 & 0.069 & 0.455 & 20.8 \\
GFIE \cite{hu2023gfie} & SD & \checkmark & 0.965 & 0.065 & 0.311 & 17.7 \\ 
GFIE \cite{hu2023gfie} + 3D & SD & \checkmark & 0.978 & \textbf{0.062} & 0.341 & 16.4 \\ 
\hline
Ours &SD &  & - & - & - & 11.1 \\
Ours* & SD & & - & - & - & 12.3 \\
Ours + GFM \cite{hu2023gfie} & SD & \checkmark & \textbf{0.987} & 0.067 & \textbf{0.260} & \textbf{10.6} \\
\hline

\end{tabular}
}
\caption{
Quantitative results on the GFIE dataset \cite{hu2023gfie}. We denote the input modalities as follows: H = head image only; SD = scene image with depth map. \textit{Ours + GFM} denotes using the gaze following modules of \cite{hu2023gfie} to refine the estimated 3D gaze direction. A check mark indicates the method requires 2D gaze following; otherwise, it directly estimates 3D gaze direction. Ours* uses a depth map from the zero-shot estimator \cite{bhat2023zoedepth}.
}
\vspace{-1.0\baselineskip}
\label{tb:gfie}
\end{center}
\end{table}

\begin{table}[t]
\begin{center}
\resizebox{\linewidth}{!}{
\begin{tabular}{lcccc}
\hline
Method & AUC $\uparrow$ & L2 Dist. $\downarrow$ & 3D Dist. $\downarrow$ & 3D MAE $\downarrow$ \\
\hline
&\multicolumn{4}{c}{w/o GFM} \\ \hline
GFIE$_\text{head}$  \cite{hu2023gfie} & - & - & - & 27.3 \\ 
Ours & - & - & - & \textbf{25.2} \\ \hline
& \multicolumn{4}{c}{w/ GFM} \\ \hline
GFIE \cite{hu2023gfie} & \textbf{0.921} & \textbf{0.114} & 0.365 & 19.8 \\ 
Ours + GFM \cite{hu2023gfie} & \textbf{0.921} & \textbf{0.114} & \textbf{0.317} & \textbf{15.8} \\
\hline
\end{tabular}
}
\caption{
Quantitative results on the CAD-120 dataset \cite{Koppula2013}. Models are trained on the GFIE dataset \cite{hu2023gfie}. GFIE$_\text{head}$ uses only the head image as input without GFM \cite{hu2023gfie}. Additional results are provided in \cref{sec:ap-cad120}.
}
\label{tb:cad120}
\end{center}
\vspace{-1.0\baselineskip}
\end{table}

\section{Experiments}
\label{sec:exp}
\paragraph{Dataset.}
This paper utilizes three publicly available benchmark datasets: the GFIE dataset \cite{hu2023gfie}, the CAD-120 dataset \cite{Koppula2013}, and the GAFA dataset \cite{nonaka2022dynamic}. The GFIE dataset \cite{hu2023gfie} includes 72K images of subjects in indoor scenes engaging in various activities, often interacting with nearby objects, such as through physical contact or proximity to objects they gaze at. Subjects are instructed to look at specified points within the scene during these activities.

The CAD-120 dataset \cite{Koppula2013}, containing 1.7K images of similar activities, is used to evaluate the generalization of models trained on the GFIE dataset to unseen scenes with different camera settings. Both datasets have visible gaze targets in the images. Following \cite{hu2023gfie}, we use real depth maps. We also test depth maps from the zero-shot metric depth estimator \cite{bhat2023zoedepth}.

To test our method in more challenging scenarios, we use the GAFA dataset \cite{nonaka2022dynamic}, which includes 882K images from four indoor and one outdoor environment. In this dataset, subjects move freely, searching for specified objects in scenes where interaction with objects is less frequent, and gaze targets may not always be visible. As it lacks depth maps, we generate them using the same zero-shot depth estimator \cite{bhat2023zoedepth}.

\paragraph{Baselines.}
In our evaluation on the GFIE dataset \cite{hu2023gfie} and the CAD-120 dataset \cite{Koppula2013}, we compare our method with the baseline approach presented in \cite{hu2023gfie}, referred to as \textit{GFIE}. First, the appearance-based estimator predicts the 3D gaze direction as a directional prior from the head image. We use the same estimator as \( f_\text{view} \) for the GFIE and the CAD-120 datasets, with its output serving as the view direction \( \mathbf{v} \). Then, a 2D gaze-following module generates a heatmap of the gazed point using RGB and depth inputs. A final non-trainable 3D module aligns this with the 3D gaze direction by computing pixel directions from the head's 3D position and selecting the one closest to the initial gaze estimate. In this paper, we abbreviate the 2D/3D gaze-following modules as \textit{GFM}. We also report other baseline results from \cite{hu2023gfie} for comparison: \textit{Random}, \textit{Center}, head-appearance-based approaches \cite{fischer2018rt, kellnhofer2019gaze360}, and 2D gaze-following approaches \cite{chong2020detecting, recasens2015they, lian2018believe}. See \cite{hu2023gfie} for details on these baselines.

For the GAFA dataset \cite{nonaka2022dynamic}, we also benchmark against the method proposed in \cite{nonaka2022dynamic}, termed GAFA$_{\text{temporal}}$. First, the appearance-based estimator predicts 3D head and body directions as directional priors from temporal frames, using a total of seven frames: the target frame, along with three future and three past frames. Each frame includes the subject's full-body RGB image, head position mask, and 2D flow of the body center. On the GAFA dataset, we use this estimator as \( f_\text{view} \),
aligning with GAFA$_{\text{temporal}}$. A temporal aggregation module then combines all frames to yield the final 3D gaze direction. We refer to \textit{GAFA} as the single-frame input version, where seven identical target frames are used as input to simulate the single-frame condition. We also include other baseline results from \cite{nonaka2022dynamic}: \textit{Fixed bias} and head-appearance-based methods \cite{kellnhofer2019gaze360, zhang2020eth}. Refer to \cite{nonaka2022dynamic} for further baseline details.

Architecture details of the appearance-based estimator for each dataset are provided in \cref{sec:ap-fview}.

To ensure a fair comparison, we report results for the baseline GFIE and GAFA with additional inputs, $P_\text{pose}$ and $P_\text{object}$, referred to as \textit{GFIE + 3D} and \textit{GAFA + 3D}, respectively. Further details are provided in \cref{sec:ap-baseline}.

\paragraph{Metrics.} 
We primarily evaluate methods using the 3D mean angular error (MAE) between predicted and ground truth 3D gaze directions. For the GFIE and the CAD-120 datasets, we additionally report AUC \cite{judd2009learning}, L2 distance, and 3D distance to measure error between predicted and ground truth gazed points based on 2D gaze-following results, when 2D gaze point estimation is available, following the previous work \cite{hu2023gfie}. For detailed metric descriptions, refer to \cite{hu2023gfie}. Note that \emph{improving 2D gaze-following performance is not the main focus} of this paper. For the GAFA dataset \cite{nonaka2022dynamic}, we also report 2D MAE, calculated using only the x and y components of the 3D gaze direction, for consistency with the previous work \cite{nonaka2022dynamic}.

\subsection{Results on the GFIE dataset \cite{hu2023gfie}}
\label{sec:gfie}
We present the quantitative performance in \cref{tb:gfie}. Without 2D gaze-following, our approach significantly outperforms all baselines in 3D MAE, achieving a 37\% improvement over the leading baseline GFIE \cite{hu2023gfie}. Ours*, which uses a depth map from a zero-shot estimator \cite{bhat2023zoedepth}, performs comparably. Additionally, when combined with the gaze-following modules, Ours + GFM \cite{hu2023gfie} outperforms GFIE across most metrics. It is important to note that the primary focus of this paper is enhancing 3D gaze direction estimation, rather than improving gaze-following. Qualitative results are shown in \cref{fig:gfie-vis}. Even in challenging scenarios where the subject's face is not visible, our method provides more accurate 3D gaze direction estimates than GFIE.

\subsection{Results on the CAD-120 dataset \cite{Koppula2013}}
We evaluated the generalization capability of models trained on the GFIE dataset \cite{hu2023gfie} by testing them on the CAD-120 dataset \cite{Koppula2013}. Quantitative results are presented in \cref{tb:cad120}. In both cases, with and without GFM, our approach outperforms the baselines, demonstrating superior generalizability in unseen scenes with different camera settings. Qualitative results are shown in \cref{fig:cad120-vis}. Even in challenging scenarios where the subject's face is unclear, our method estimates 3D gaze more accurately than GFIE. Results with the other baselines are discussed in \cref{sec:ap-cad120}.

\begin{figure}[t]
    \begin{center}
    \includegraphics[width=\linewidth]{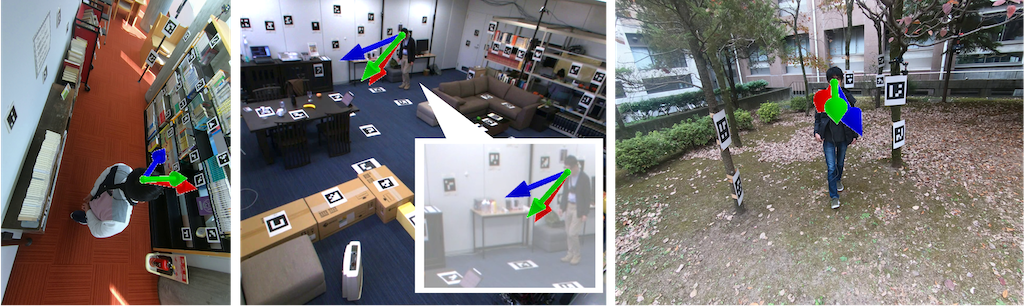}
    \end{center}
   \vspace{-1.5\baselineskip}
    \caption{Qualitative results on the GAFA dataset \cite{nonaka2022dynamic}. Red, green, and blue arrows indicate the ground truth, Ours, and GAFA \cite{nonaka2022dynamic}. }
    \label{fig:gafa-vis}
    \vspace{-.5\baselineskip}
\end{figure}

\begin{table}[t]
\centering
\resizebox{0.95\columnwidth}{!}{%
\begin{tabular}{lccccccc}
\hline
Method & Input & Office & LR & Kitchen & Library & Courtyard & All \\
\hline
Fixed bias & & 88.0/76.0 & 85.5/76.7 & 86.0/82.4 & 89.0/85.1 & 89.7/88.7 & 88.1/79.7 \\
Gaze360 \cite{kellnhofer2019gaze360}& SBI   & 24.0/19.2 & 41.1/31.3 & 32.4/21.2 & 27.5/20.7 & 28.2/28.3 & 30.4/24.5 \\
XGaze \cite{zhang2020eth} & SBI  & 24.2/23.0 & 42.0/40.9 & 23.3/22.9 & 24.6/22.3 & 30.2/31.9 & 29.2/28.4 \\
GAFA$_\text{temporal}$  \cite{nonaka2022dynamic} & TBI & 14.4/14.3 & 25.1/22.6 & 20.4/19.6 & 19.8/18.4 & 25.4/26.9 & 21.7/20.9 \\
GAFA \cite{nonaka2022dynamic} & SBI & 16.1/15.8 & 26.0/23.2 & 21.1/20.5 & 20.9/19.5 & 27.2/28.5 & 22.9/22.1 \\
GAFA \cite{nonaka2022dynamic} + 3D & SSD & 20.5/20.3 & \textbf{20.6}/\textbf{19.3} & \textbf{16.3}/\textbf{16.3}  & 28.5/29.2 & \textbf{24.4}/\textbf{23.5} & 22.8/22.4 \\
\hline
Ours & SSD & \textbf{13.5/}\textbf{13.1} & 21.9/23.9 & 16.9/16.9 & \textbf{17.3}/\textbf{16.3} & 26.2/28.6 & \textbf{19.9}/\textbf{20.6} \\
\hline
\end{tabular}%
}
\caption{
Quantitative results on the GAFA dataset \cite{nonaka2022dynamic}. Input modalities are denoted as follows: the first letter indicates frame type (S = single; T = temporal), the second denotes image type (H = head; B = full-body; S = scene), and the third represents modality (I = image-only; D = depth features).
}
\label{tb:gafa}
\end{table}

\subsection{Results on the GAFA dataset \cite{nonaka2022dynamic}}
\label{sec:gafa}
The quantitative performance is presented in \cref{tb:gafa}, where our method consistently outperforms all baselines, achieving an average improvement of 13\% in 3D MAE and 7\% in 2D MAE over the leading baseline GAFA \cite{nonaka2022dynamic}, in the single-frame setting. Qualitative results are shown in \cref{fig:gafa-vis}, demonstrating that our approach estimates 3D gaze direction more accurately than GAFA in challenging scenarios, including backward-facing subjects, subjects far from the cameras, extreme camera angles, and cases where the gazed target is outside the camera's field of view.

\subsection{Ablation studies}
\label{sec:abl}
\begin{table}[t]
\begin{center}
\resizebox{\columnwidth}{!}{
\begin{tabular}{lcc}
\hline
Method & GFIE \cite{hu2023gfie}  & GAFA \cite{nonaka2022dynamic} \\ \hline
Appearance & 19.4 &22.9 \\
Appearance + Pose & 13.1 &  20.3 \\
Appearance + Pose + Object & \textbf{11.1} &  \textbf{19.9} \\
\hline
\end{tabular}
}
\caption{Quantitative results on the effect of 3D pose and object positions in 3D MAE.}
\label{tb:abl-pose-obj}
\end{center}
\vspace{-1.0\baselineskip}
\end{table}

\paragraph{Effect of 3D understanding of pose and objects.}
We evaluate the impact of incorporating 3D understanding of both pose and objects, as shown quantitatively in \cref{tb:abl-pose-obj}. \textit{Appearance} refers to models that rely solely on the subject's appearance as input, without incorporating 3D context. Model details are provided in \cref{sec:ap-abl-pose-obj-mod}. First, we incorporate 3D pose as an additional context cue for the subject, following a setup similar to GAFA \cite{nonaka2022dynamic}. We then include object positions as a scene-level context cue, complementing the subject's context cues, inspired by GFIE \cite{hu2023gfie}.

Results for the model that combines appearance with 3D pose $P_\text{pose}$, denoted as \textit{Appearance + Pose} in \cref{tb:abl-pose-obj}, show performance improvements in both the GFIE and the GAFA datasets. To isolate the effect of object features, we use a constant vector as the source sequence for $f_\text{decoder}$. The \textit{Appearance + Pose + Object} configuration, which incorporates both 3D pose and object positions, yields even better accuracy, particularly on the GFIE dataset compared to the GAFA dataset. This difference may stem from the GAFA dataset's challenging scenarios where subjects interact less with objects than in the GFIE dataset, yet the performance improvements are still observed by considering objects. 

\begin{figure}[t]
    \begin{center}
    \includegraphics[width=\columnwidth]{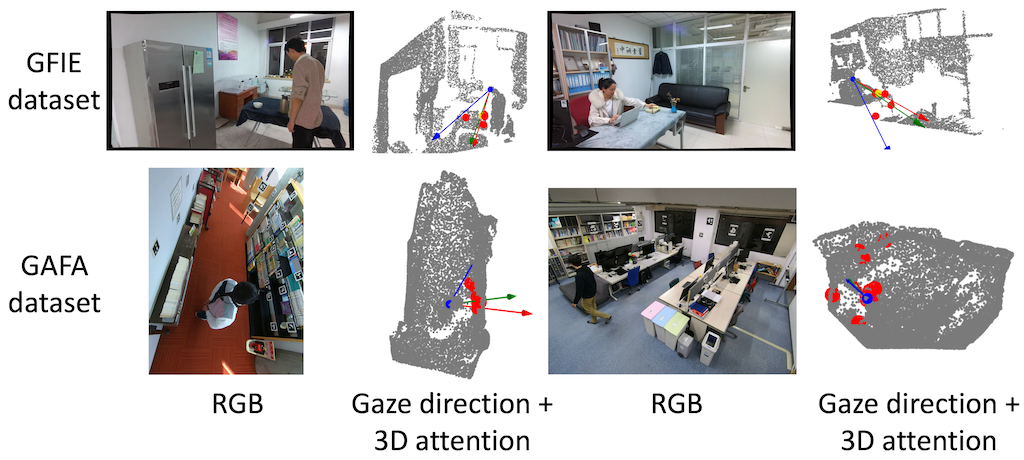}
    \end{center}
   \vspace{-1.5\baselineskip}
    \caption{Visualization of decoder attention to object positions $P_\text{object}$. In each pair, left image shows the input RGB image, and the right image shows the object positions with strong attention values and overlaid gaze directions. A yellow dot indicates the ground truth 3D gazed point in the GFIE dataset \cite{hu2023gfie}. Red, green, and blue arrows indicate the 3D gaze direction of the ground truth, Ours, and the baseline GFIE for the GFIE dataset \cite{hu2023gfie} and the baseline GAFA for the GAFA dataset \cite{nonaka2022dynamic}, respectively.}
     \vspace{-.5\baselineskip}
    \label{fig:attn-vis}
\end{figure}

Performance gains are especially pronounced when the subject is closer to the gazed objects, as illustrated in the leftmost image of \cref{fig:gafa-vis}. We visualize the decoder's attention to object positions $P_\text{object}$ in \cref{fig:attn-vis}. The leftmost figure in each triplet highlights object positions with strong attention values, determined by the 95th percentile threshold. In the GFIE dataset, objects close to the ground-truth 3D gaze point (colored yellow) receive high attention. In the GAFA dataset \cite{nonaka2022dynamic}, while more diverse than the GFIE dataset \cite{hu2023gfie}, object positions around the ground-truth gaze direction also exhibit strong attention, suggesting that nearby object positions significantly impact the final gaze estimation.

\begin{table}[t]
\begin{center}
\resizebox{\columnwidth}{!}{
\begin{tabular}{lcc}
\hline
Method & GFIE \cite{hu2023gfie} & GAFA \cite{nonaka2022dynamic} \\ \hline
Appearance & 19.4 & 22.9 \\
Appearance + 3D & 18.9 & 21.2  \\
Appearance + 3D + GA3CE & \textbf{11.1} &  \textbf{19.9} \\
\hline
No egocentric transformation & 14.2 & 20.8 \\
No cyclotorsion rotation & 12.1 & 20.1  \\ 
No D$^3$ positional encoding & 12.9 & 20.9 \\
All & \textbf{11.1} & \textbf{19.9} \\
\hline
\end{tabular}
}
\caption{
Quantitative results on the impact of GA3CE in 3D MAE. In the top rows, \textit{3D} indicates using pose and objects as 3D context input besides view direction. In the bottom rows, the proposed techniques in GA3CE is disabled one at a time, and \textit{All} means enabling all of them.
}
\label{tb:abl-subj-cent}
\end{center}
\vspace{-1.0\baselineskip}
\end{table}

\paragraph{Effect of 3D context representations.}
We analyze the impact of GA3CE. Quantitative results are presented in \cref{tb:abl-subj-cent}. The top rows demonstrate the influence of GA3CE, showing that simply adding 3D representations without the proposed context encoding does not achieve optimal performance. Similar findings are observed for the baseline methods, GFIE + 3D and GAFA + 3D, in \cref{tb:gfie} and \cref{tb:gafa}, where adding 3D representation alone yields only modest improvement. The bottom rows of \cref{tb:abl-subj-cent} provide further analysis, showing that disabling each component reduces performance, while enabling all components results in the best performance.

\section{Conclusion}
We propose a novel 3D gaze estimation framework that models the spatial relationship between a subject's 3D pose and object positions in the scene. Central to our approach is GA3CE, gaze-aware 3D context encoding with two key components: egocentric transformation that normalizes 3D context input to a subject-centric space, and D$^3$ positional encoding that effectively captures the directional and distance relationships between 3D context and 3D gaze. We show that GA3CE substantially improves reasoning about gaze direction in 3D space.
Extensive evaluations on three benchmark datasets demonstrate that our method consistently outperforms leading baselines in challenging scenarios.
    
\newpage
\bibliographystyle{splncs04}
\bibliography{main}
\clearpage %

\twocolumn[{%
 \centering
 \textbf{\Large  Supplementary Material }
 \vspace{20pt}
}]

\appendix
\section{Egocentrict transformation}
\subsection{Pose normalization}
\label{sec:ap-pose}
In this section, we describe the computation of \(\mathbf{t}_\text{pose}\) and \(s\), which are used to normalize the 3D pose \(P_\text{pose}\) in the egocentric transformation introduced in \cref{sec:sub-cent}. For 3D pose estimation, we use the off-the-shelf estimator from \cite{Sarandi2023dozens}, which outputs 3D poses in the \textit{smpl+head\_30} format, comprising 30 keypoints representing the body and head.

The head position, \(\mathbf{t}_\text{pose}\), is defined as the average 3D position of the head keypoints (indices 24 to 29). The scale \(s\), representing the inverse head size, is computed as the reciprocal of the L2 norm between the 24th and 28th keypoints, which defines the head's width.

To simplify the representation, we subsample the keypoints using the indices \(\{0, 4\text{--}8, 12, 18\text{--}21, 24, 26, 28, 29\}\), resulting in a reduced set of \(N_\text{pose} = 15\) keypoints, as defined in \cref{sec:contextRepr}.
  
\subsection{Details on rotation alignment}
\label{sec:ap-rot}
Human vision studies \cite{ma2015learning,einhauser2008objects,findlay2003active} have shown that gaze fixations often concentrate near the center of the visual field, with distance information playing a critical role in gaze saliency prediction, particularly from a first-person perspective. This suggests that both the direction and distance of objects relative to the subject are key factors in estimating gaze direction. For example, individuals tend to focus more on nearby objects than distant ones, even when both lie in the same direction. Likewise, objects near the visual center attract more attention than those in the periphery, even if the latter are physically closer.

Furthermore, learning spatial relationships in a normalized space helps simplify the complex patterns of poses, object positions, and their interrelations. To this end, we normalize 2D observations into an egocentric view as the 3D context representations, as described in \cref{sec:contextRepr}.

We assume that surveillance or monitoring cameras have their x-axis aligned with the horizon, the y-axis pointing downward (though not necessarily perpendicular), and the z-axis extending forward from the camera center, consistent with typical egocentric human perspectives. This assumption generally holds for common 2-DoF fixed cameras mounted on flat ceilings or walls. Even if the camera is tilted due to roll (rotation about the z-axis), this can be corrected using calibrated camera poses or scene cues such as vanishing points \cite{lee2015real,park_localization_vanishing_point}.

As a result, any rotation about the camera's z-axis within the camera coordinate system, which rotates the x-y plane of the view, disrupts the consistent alignment of the camera view that maintains the horizon in the view as horizontal.
Our goal is to normalize 3D context representations to this aligned view. During egocentric transformation, we apply rotation that maintain the consistency of the x-y plane w.r.t. the horizon, by avoiding rotations around the z-axis.

\begin{figure}[t]
    \begin{center}
    \includegraphics[width=0.7\columnwidth]{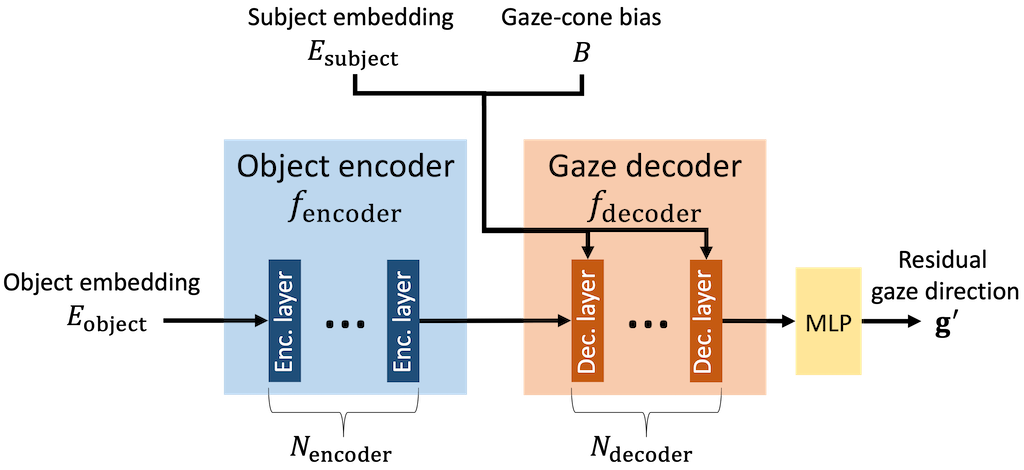}
    \end{center}
   \vspace{-1.0\baselineskip}
\caption{Architecture of the 3D gaze transformer. \textit{Enc. layer} and \textit{Dec. layer} refer to the transformer encoder and decoder layers, respectively.
}
    \label{fig:gazetransformer}
\end{figure}

\begin{table}[t]
\begin{center}
\resizebox{\columnwidth}{!}{
\begin{tabular}{lcccccc}
\hline
Method & 2D gaze &  AUC $\uparrow$ & L2 Dist. $\downarrow$ & 3D Dist. $\downarrow$ & 3D MAE $\downarrow$ \\
\hline
GFIE \cite{hu2023gfie} & \checkmark & 0.965 & 0.065 & 0.311 & 17.7 \\ 
GFIE \cite{hu2023gfie} + 3D & \checkmark & 0.978 & 0.062 & 0.341 & 16.4 \\ 
GFIE \cite{hu2023gfie} + ViTGaze \cite{song2024vitgaze} & \checkmark & 0.965 & \textbf{0.054} & 0.32 & 17.9 \\ 
GFIE \cite{hu2023gfie} + 3D + ViTGaze \cite{song2024vitgaze} & \checkmark & 0.978 & \textbf{0.054} & 0.30 & 16.6 \\ 
\hline
Ours  &  & - & - & - & 11.1 \\
Ours + GFM \cite{hu2023gfie} & \checkmark & \textbf{0.987} & 0.067 & \textbf{0.260} & \textbf{10.6} \\
\hline
\end{tabular}
}

\caption{
Quantitative results on the GFIE dataset \cite{hu2023gfie}. \textit{ViTGaze} refers to replacing the baseline GFIE's 2D gaze-following module from \cite{song2024vitgaze}.
}
\label{tb:ap-vitgaze}
\end{center}
\end{table}

\begin{figure*}[t]
    \begin{center}
    \includegraphics[width=\linewidth]{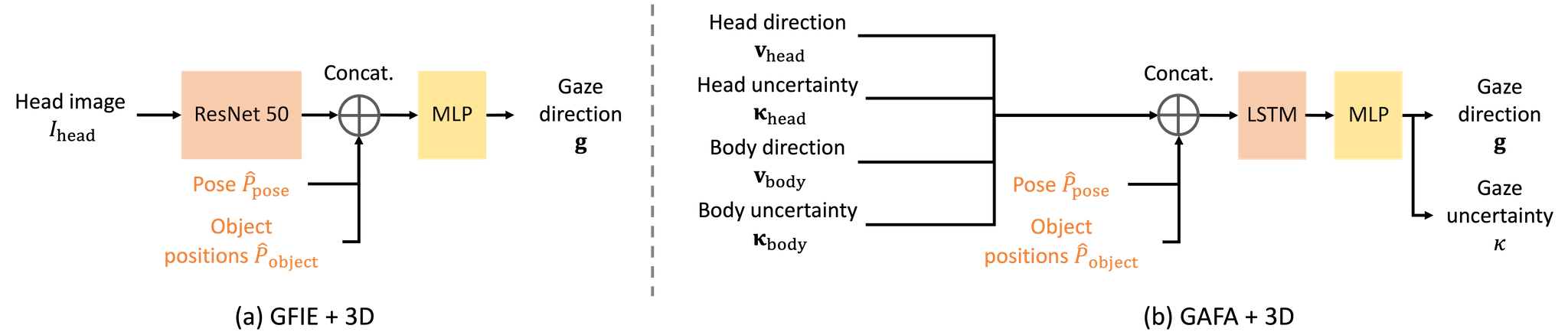}
    \end{center}
   \vspace{-1.0\baselineskip}
    \caption{
Illustration of the modifications to the baseline architectures for incorporating 3D context in (a) GFIE \cite{hu2023gfie} and (b) GAFA \cite{nonaka2022dynamic}.
    }
    \label{fig:gfie_gafa_modify}
\end{figure*}

\begin{figure*}[t]
    \begin{center}
    \includegraphics[width=\linewidth]{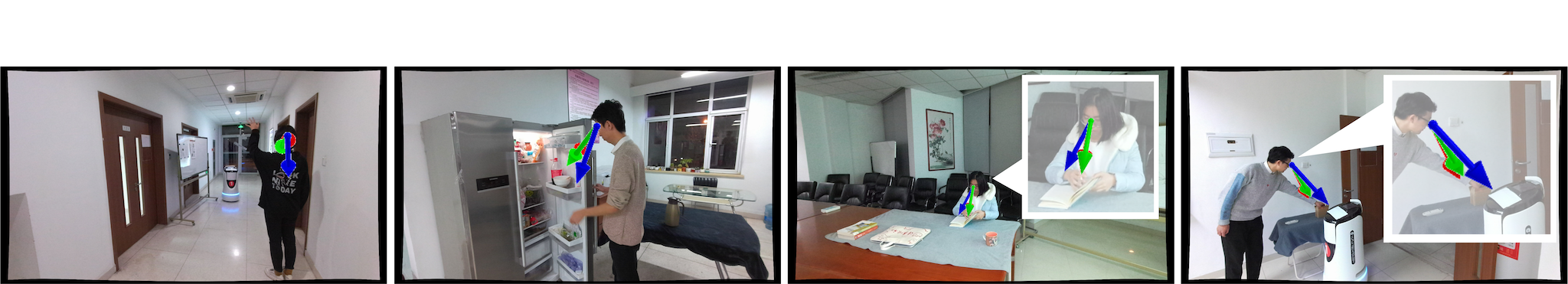}
    \end{center}
   \vspace{-1.0\baselineskip}
    \caption{
Additional qualitative results on the GFIE dataset \cite{hu2023gfie}. Red, green, and blue arrows indicate the ground truth, Ours, and the baseline GFIE \cite{hu2023gfie}, respectively.
    }
    \label{fig:supp-gfie-vis}
\end{figure*}

\begin{table}[t]
\begin{center}
\resizebox{\columnwidth}{!}{
\begin{tabular}{lcccccc}
\hline
Method & Input & 2D gaze &  AUC $\uparrow$ & L2 Dist. $\downarrow$ & 3D Dist. $\downarrow$ & 3D MAE $\downarrow$ \\
\hline
Random & & & 0.469 & 0.758 & 1.910 & 70.3 \\
Center & & & 0.456 & 0.706 & 1.280 & 75.9 \\
Rt-Gene \cite{fischer2018rt} & H & &  0.463 & 0.492 & 0.483 & 26.5 \\
Gaze360 \cite{kellnhofer2019gaze360} & H &  & 0.463 & 0.474 & 0.427 & 20.6 \\
GazeFollow \cite{recasens2015they}  & SD & \checkmark &  0.862 & 0.196 & 1.030 & 44.1 \\
Lian \cite{lian2018believe} & SD &\checkmark & 0.871 & 0.180 & 0.813 & 34.8 \\
Chong \cite{chong2020detecting}  & SD &\checkmark &  0.891 & 0.152 & 0.812 & 31.9 \\
GFIE$_\text{head}$  \cite{hu2023gfie}& H & & - & - & - & 27.3 \\ 
GFIE \cite{hu2023gfie}& SD & \checkmark & 0.921 & 0.114 & 0.365 & 19.8 \\
GFIE \cite{hu2023gfie} + 3D & SD & \checkmark  & \textbf{0.933} & \textbf{0.094} & 0.365 & 17.8 \\ \hline
Ours*& SD &  & - & - & - & 24.5 \\
Ours& SD &  & - & - & - & 25.2 \\
Ours + GFM \cite{hu2023gfie}& SD & \checkmark & 0.921 & 0.094 & 0.314 & 15.8 \\
Ours + GFM\textdagger \cite{hu2023gfie}& SD & \checkmark & \textbf{0.933} & \textbf{0.094} & \textbf{0.243} & \textbf{14.6} \\
\hline
\end{tabular}
}
\caption{
Quantitative results on the CAD-120 dataset \cite{Koppula2013}, with baselines from \cite{hu2023gfie}, are presented. Input modalities are denoted as follows: H = head image only; SD = scene image with depth map. A check mark indicates that the method requires 2D gaze following; otherwise, it directly estimates the 3D gaze direction. \textit{GFM} refers to the gaze following modules from \cite{hu2023gfie}. GFIE$_\text{head}$ uses only the head image as input without GFM \cite{hu2023gfie}. Ours* uses a depth map from the zero-shot estimator \cite{bhat2023zoedepth}. GFM\textdagger$\,$ uses the same module trained for GFIE + 3D, as detailed in \cref{sec:modify-baseline}.
} 
\label{tb:cad120_full}
\end{center}
\end{table}

\begin{figure*}[t]
    \vspace{-1.5\baselineskip}
    \begin{center}
    \includegraphics[width=\linewidth]{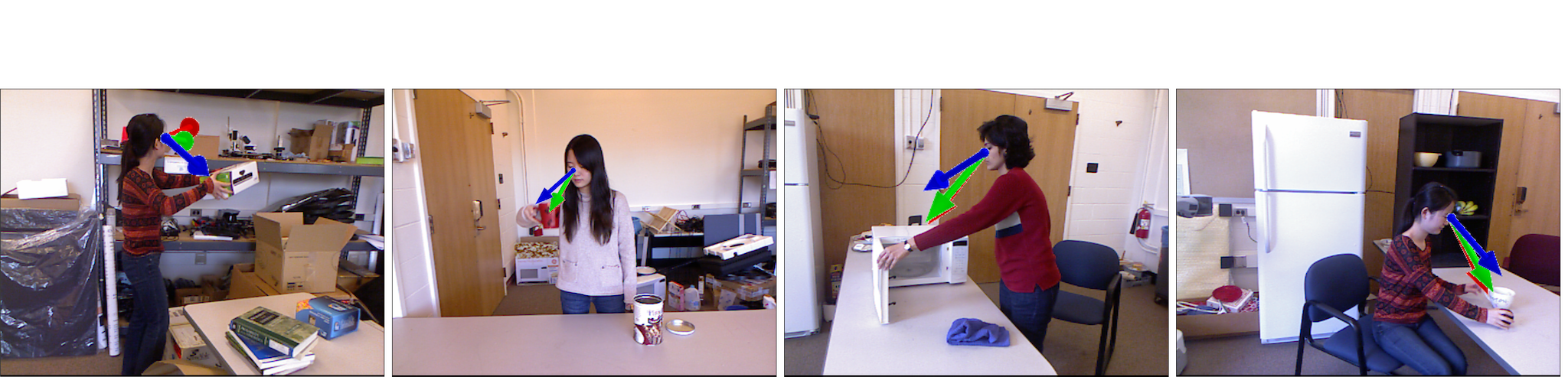}
    \end{center}
   \vspace{-1.0\baselineskip}
   \caption{Additional qualitative results on the CAD-120 dataset \cite{Koppula2013}. Red, green, and blue arrows represent the ground truth, Ours + GFM \cite{hu2023gfie}, and the baseline GFIE \cite{hu2023gfie}, respectively.}  
    \label{fig:supp-cad120-vis}
\end{figure*}

\begin{figure*}[t]
    \begin{center}
    \includegraphics[width=\linewidth]{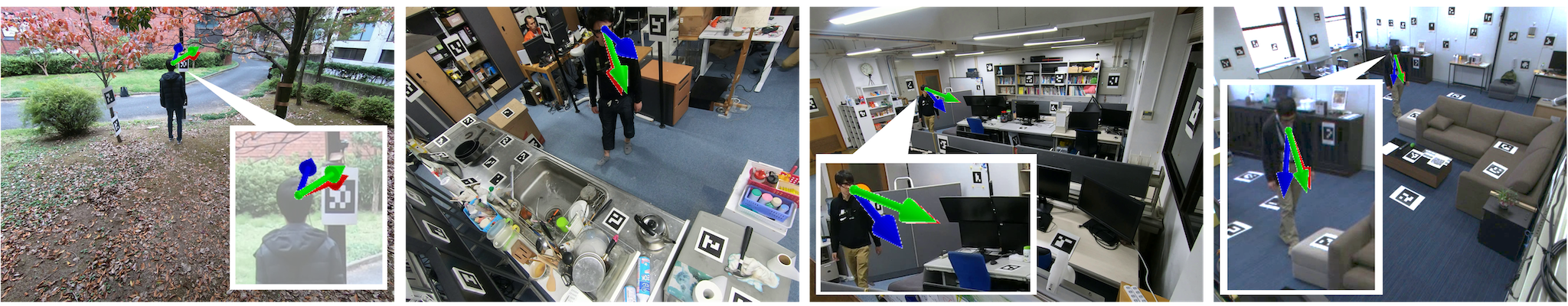}
    \end{center}
   \vspace{-1.0\baselineskip}
   \caption{Additional qualitative results on the GAFA dataset \cite{nonaka2022dynamic}. Red, green, and blue arrows represent the ground truth, Ours, and the baseline GAFA \cite{nonaka2022dynamic}, respectively.}  
    \label{fig:supp-gafa-vis}
\end{figure*}

\begin{figure*}[t]
    \begin{center}
    \includegraphics[width=\linewidth]{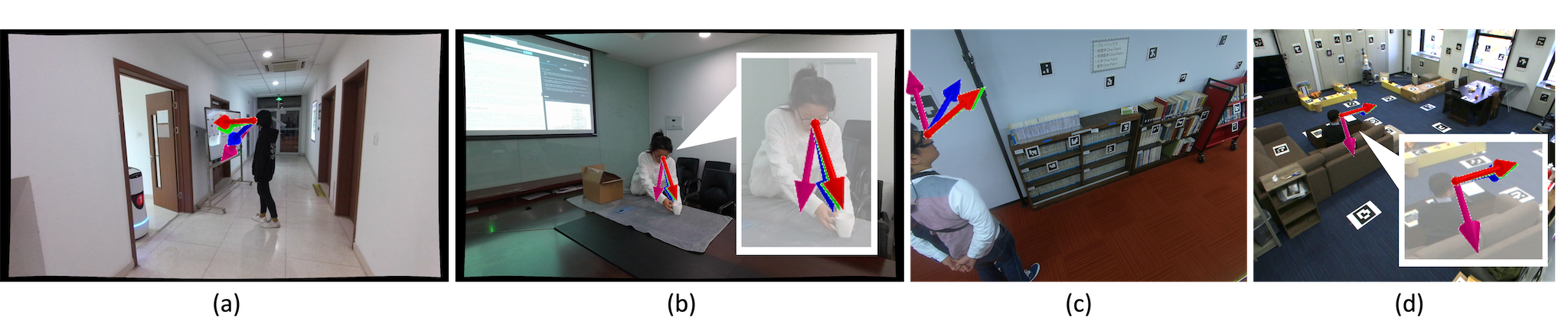}
    \end{center}
   \vspace{-1.0\baselineskip}
   \caption{
Qualitative ablation results illustrating the effects of pose and object understanding. The red arrow indicates the ground truth, while the magenta, blue, and green arrows represent predictions from the \textit{Appearance}, \textit{Appearance + Pose}, and \textit{Appearance + Pose + Object} models, respectively, as described in \cref{tb:abl-pose-obj} of the main paper. (a) and (b) present results from the GFIE dataset \cite{hu2023gfie}, while (c) and (d) show results from the GAFA dataset \cite{nonaka2022dynamic}.
   }
    \label{fig:supp-pose-obj-abl}
\end{figure*}

\begin{figure*}[t]
    \begin{center}
    \includegraphics[width=\linewidth]{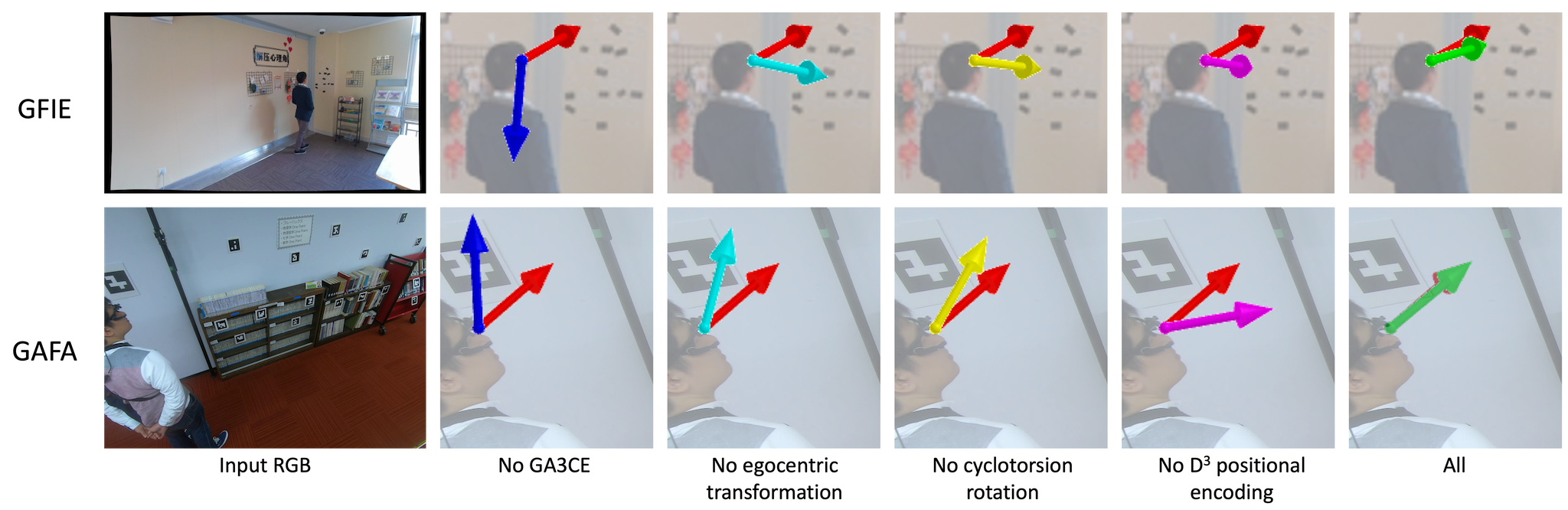}
    \end{center}
   \vspace{-1.0\baselineskip}
   \caption{
Qualitative ablation results for gaze-aware 3D context encoding and the proposed components. The top row shows results on the GFIE \cite{hu2023gfie} dataset, while the bottom row presents results on the GAFA \cite{nonaka2022dynamic} dataset. The red arrow indicates the ground truth, and the other arrows represent outputs from ablated models. \textit{All} denotes the full model with all proposed components.
   }  
    \label{fig:supp-ga3fe-abl}
\end{figure*}

\begin{figure*}[t]
    \begin{center}
    \includegraphics[width=\linewidth]{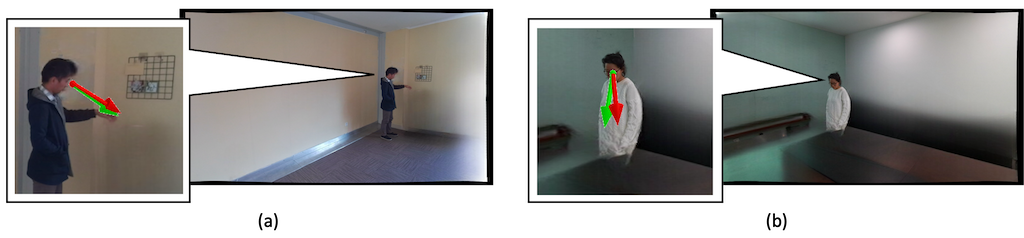}
    \end{center}
   \vspace{-1.0\baselineskip}
    \caption{
Visualization of cases where objects are missing along the gaze direction. Red and green arrows indicate the ground truth and the prediction, respectively. In (a), objects lie outside the gaze direction. In (b), no objects are present near the gaze direction.
    }
    \label{fig:ap-flat-surface}
\end{figure*}

\begin{figure*}[t]
    \begin{center}
    \includegraphics[width=\linewidth]{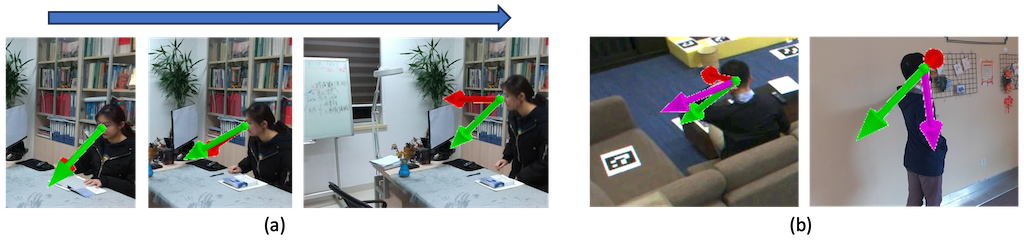}
    \end{center}
   \vspace{-1.0\baselineskip}
    \caption{
Visualization of failure cases. Red and green arrows denote the ground truth and the predictions, respectively. In (a), the blue arrow indicates the temporal direction. In (b), the magenta arrow represents the view direction $\mathbf{v}$.
    }
    \label{fig:ap-failure-cases}
\end{figure*}

\subsection{Axis-angle rotation alignment}
\label{sec:ap-rot-aa}
An axis-angle representation is defined by a rotation axis $\mathbf{a} \in \mathbb{S}^2$ and a rotation angle $\eta \in \mathbb{R}$. To align the view direction $\mathbf{v} \in \mathbb{S}^2$ with a fixed direction $\mathbf{z} = [0, 0, 1]$, the axis $\mathbf{a}$ and angle $\eta$ are given by:

\begin{equation}
\label{eq:ap-aa}
\begin{split}
    \mathbf{a} &= \frac{\mathbf{v} \times \mathbf{z}}{\Vert \mathbf{v} \times \mathbf{z} \Vert}, \\
    \eta &= \arccos(\mathbf{v}^T \mathbf{z}).
\end{split}
\end{equation}

To examine how the rotation in \cref{eq:ap-aa} affects rotation around the z-axis, we convert it to intrinsic Euler angles in x-y-z order, denoted as $\{\theta, \phi, \psi\}$. The corresponding rotation matrix is expressed as $R_\text{euler} = R_x(\theta) R_y(\phi) R_z(\psi)$.

Using Rodrigues' rotation formula, the Euler angle $\psi$, representing the rotation around the z-axis, is computed as:

\begin{equation}
    \psi = \text{arctan2}(c_\eta + a_x^2(1 - c_\eta), a_x a_y(1 - c_\eta) - a_z s_\eta),
\end{equation}

where $\mathbf{a} = [a_x, a_y, a_z]$, $c_\eta = \cos \eta$, and $s_\eta = \sin \eta$. Unless $a_x a_y(1 - c_\eta) - a_z s_\eta$ equals zero, the rotation defined by \cref{eq:ap-aa} includes a nonzero component around the z-axis.

\subsection{Analytical solution for cyclotorsion rotation}
\label{sec:ap-rot-cr}
Our objective is to keep $\psi$ at zero and determine the angles $\theta$ and $\phi$ that satisfy the constraint on the view direction $\mathbf{v} = [v_x, v_y, v_z]$ relative to $\mathbf{z} = [0, 0, 1]$, as described in \cref{eq:in-plane}:
\begin{equation}
\begin{bmatrix}
0 \\
0 \\
1 \\
\end{bmatrix} = 
\begin{bmatrix}
\cos\phi & 0 & \sin\phi \\
\sin\theta\sin\phi & \cos\theta & -\sin\theta\cos\phi \\
-\cos\theta\sin\phi & \sin\theta & \cos\theta\cos\phi \\
\end{bmatrix}
\begin{bmatrix}
v_x \\
v_y \\
v_z \\
\end{bmatrix}.
\end{equation}
Then, $\phi$ is defined as:
\begin{equation}
\phi = \text{arctan2}(-v_x, v_z)
\end{equation}
Given $\phi$, $\theta$ is defined as:
\begin{equation}
    \theta = \text{arctan2}(v_y, v_z\cos\phi - v_x\sin\phi)
\end{equation}

\section{Additional implementation details}
\subsection{Details on positional encoding}
\label{sec:supp-pe}
We present the formulation and implementation details of the positional encoding in this section.

Since positional encoding is applied independently to each point in the normalized body keypoints \( P'_\text{pose} \) and normalized object positions \( P'_\text{object} \), we omit \( N_\text{pose} \) and \( N_\text{object} \) in the following formulations for simplicity.

The standard positional encoding \cite{tancik2020fourier} for the view direction \( \mathbf{v}' \) is defined as:
\begin{equation}
    \gamma_\text{view}: \mathbb{S}^2 \to \mathbb{R}^{C_\text{gaze}}.
\end{equation}

The D$^3$ positional encoding \( \tilde{\gamma}_\text{pose} \) for a point in \( P'_\text{pose} \), along with the standard positional encodings \( \gamma_\text{pose}^\text{dir} \) and \( \gamma_\text{pose}^\text{dist} \) for its direction and distance components, are defined as:
\begin{equation}
\begin{split}
    \tilde{\gamma}_\text{pose}: \mathbb{R}^3 &\to \mathbb{R}^{C_\text{keypoint}} \\
    \gamma_\text{pose}^\text{dir}: \mathbb{S}^2 &\to \mathbb{R}^{C_\text{keypoint}^\text{dir}} \\
    \gamma_\text{pose}^\text{dist}: \mathbb{R} &\to \mathbb{R}^{C_\text{keypoint}^\text{dist}}
\end{split}
\end{equation}
where \( C_\text{keypoint}^\text{dir} = 6 \), \( C_\text{keypoint}^\text{dist} = 4 \), and \( C_\text{keypoint} = C_\text{keypoint}^\text{dir} + C_\text{keypoint}^\text{dist} = 10 \).

Similarly, the D$^3$ positional encoding \( \tilde{\gamma}_\text{object} \) for a point in \( P'_\text{object} \), along with the standard positional encodings \( \gamma_\text{object}^\text{dir} \) and \( \gamma_\text{object}^\text{dist} \) for direction and distance, are defined as:
\begin{equation}
\begin{split}
    \tilde{\gamma}_\text{object}: \mathbb{R}^3 &\to \mathbb{R}^{C_\text{latent}} \\
    \gamma_\text{object}^\text{dir}: \mathbb{S}^2 &\to \mathbb{R}^{C_\text{latent}^\text{dir}} \\
    \gamma_\text{object}^\text{dist}: \mathbb{R} &\to \mathbb{R}^{C_\text{latent}^\text{dist}}
\end{split}
\end{equation}
where \( C_\text{latent}^\text{dir} = 128 \), \( C_\text{latent}^\text{dist} = 128 \), and \( C_\text{latent} = C_\text{latent}^\text{dir} + C_\text{latent}^\text{dist} = 256 \).

\subsection{Architecture details of 3D gaze transformer}
\label{sec:ap-transformer}
The architecture is illustrated in \cref{fig:gazetransformer}. We use the transformer module \cite{vaswani2017attention} from PyTorch \cite{paszke2019pytorch}. The object encoder \( f_\text{encoder} \) and the gaze decoder \( f_\text{decoder} \) consist of \( N_\text{encoder} = 3 \) transformer encoder layers and \( N_\text{decoder} = 3 \) transformer decoder layers, respectively. The feedforward network has a dimension of 512, and the multi-head attention \cite{vaswani2017attention} uses 2 heads. Other hyperparameters follow the default settings in \cite{paszke2019pytorch}. The transformer processes the object feature \( E_\text{object} \) as the source sequence and the subject feature \( E_\text{subject} \) as the target sequence. 

Following \cite{tonini2023object}, we incorporate a gaze-cone-based additive attention bias \( B \in \mathbb{R}^{N_\text{object}} \) to the object features \( F_\text{object} \) in the cross-attention between \( E_\text{subject} \) and \( F_\text{object} \). This bias emphasizes object features aligned with the subject's view direction \( \mathbf{v} \), where each element of the bias is defined as the cosine similarity between \( \mathbf{v} \) and \( \mathbf{p}_\text{object} \in P_\text{object} \).

For batch processing in the attention layers, the number of object positions is padded to a maximum \( N_\text{object}^\text{max} \geq N_\text{object} \). Specifically, \( N_\text{object}^\text{max} \) is set to 168 for the GFIE \cite{hu2023gfie} and CAD-120 \cite{Koppula2013} datasets, and 278 for the GAFA dataset \cite{nonaka2022dynamic}. During multi-head attention, non-existent object positions are masked out.

Finally, the residual gaze direction \( \mathbf{g}^\prime \) is decoded using a two-layer MLP with 512 hidden units and ReLU activation.

\label{sec:modify-baseline}  

\subsection{Training details}
\label{sec:ap-training}
The batch size is set to 32 for the GFIE dataset \cite{hu2023gfie} and 64 for the GAFA dataset \cite{nonaka2022dynamic} to accommodate the larger dataset size and improve training efficiency. The network is trained for 20 epochs on a single A10 GPU using the AdamW optimizer \cite{loshchilov2017decoupled} with a learning rate of 0.0014. Cosine scheduling with a 4-epoch warm-up is employed. Weight decay is set to 0.1, and gradient clipping with an L2-norm threshold of 0.1 is applied.

For the GFIE dataset, noise $\mathbf{\epsilon}$ is added to the view direction $\mathbf{v}$ as an augmentation to mitigate overfitting. The perturbed view direction $\mathbf{v}_\text{noise}$ is computed as $\mathbf{v}_\text{noise} = \frac{\mathbf{v} + \mathbf{\epsilon}}{\Vert \mathbf{v} + \mathbf{\epsilon} \Vert}$, where $\mathbf{\epsilon} \in [-0.5, 0.5]^3$. This results in an average angular shift of approximately 22 degrees. For the GAFA dataset, however, this augmentation increased validation error and was therefore only applied during training on the GFIE dataset.

\subsection{Architecture details of the appearance-based estimators}
\label{sec:ap-fview}  
The appearance-based estimator in the baseline GFIE \cite{hu2023gfie} uses a ResNet50 image encoder followed by a gaze prediction head composed of MLPs. It takes an RGB head image as input and outputs a 3D gaze direction represented as a unit vector.

In the baseline GAFA \cite{nonaka2022dynamic}, the appearance-based estimator corresponds to the \textit{Head and Body Network} described in \cite{nonaka2022dynamic}. It processes seven temporal frames: the target frame, along with three future and three past frames. Each frame includes a full-body RGB image, a 2D head position mask, and the 2D velocity of the body center. The full-body image and head position mask are encoded separately using 2D convolutional networks, while the body velocity is encoded using MLPs. These three features are concatenated and passed through LSTM layers, which predict the directions and uncertainties of both the body and head for each frame, modeled as parameters of a 3D von Mises-Fisher distribution.

\subsection{Gaze-following modules (GFM) \cite{hu2023gfie}}
GFM, as referenced in \cref{tb:gfie,tb:cad120_full}, refers to the 2D/3D gaze-following modules from \cite{hu2023gfie}. The 2D module is a ResNet50-based convolutional autoencoder that takes as input a multi-channel 2D feature comprising a scene RGB image, a head position mask, and a field-of-view (FoV) feature map. It outputs a 2D heatmap indicating the gazed point. The FoV feature map encodes the pixel-wise directional similarity between the estimated gaze direction (obtained from the head image, as described in \cref{sec:ap-fview}) and each 3D point backprojected from the depth map using the provided camera intrinsics. All 3D points are normalized relative to the known 3D head position.

The 3D gaze-following module takes the 2D heatmap and estimated gaze direction as input and outputs the 3D gazed point. This module is non-learnable and deterministic: it selects the backprojected 3D point corresponding to the pixel location with the highest similarity in the FoV feature map near the peak of the 2D heatmap. The final 3D gaze direction is then computed as the vector from the known head position to the estimated 3D point.

\subsection{Baseline modification for 3D context input}
\label{sec:ap-baseline}
As described in \cref{sec:exp}, we modified the baseline methods GFIE \cite{hu2023gfie} and GAFA \cite{nonaka2022dynamic} to incorporate 3D pose $P_\text{pose}$ and object positions $P_\text{object}$, aligning their inputs with our approach for the corresponding datasets. Specifically, the pose $P_\text{pose}$ is normalized using the head position $\mathbf{t}_\text{head}$ and head size $s$, resulting in $\hat{P}_\text{pose}$. Similarly, object positions $P_\text{object}$ are normalized by the head position $\mathbf{t}_\text{object}$ and scaled so that their largest extent equals one, yielding $\hat{P}_\text{object}$.

These modifications are illustrated in \cref{fig:gfie_gafa_modify}, which highlights the updated modules in the pipeline while omitting other baseline components for clarity. Full pipeline details are available in \cite{hu2023gfie} and \cite{nonaka2022dynamic}. As noted in \cref{sec:exp}, we refer to the modified methods as \textit{GFIE + 3D} and \textit{GAFA + 3D}.

GFIE + 3D was trained using the publicly available code from \cite{hu2023gfie}. In the case of GAFA \cite{nonaka2022dynamic}, training the entire pipeline, including Head and Body Network along with the gaze estimation module, as shown in \cref{fig:gfie_gafa_modify} (b), led to overfitting without convergence. Improved results were obtained by training only the gaze estimation module while keeping the pre-trained weights for Head and Body Network frozen. For all other training settings, refer to \cite{nonaka2022dynamic}.

\subsection{Pipeline modification for ablation study on pose and object}
\label{sec:ap-abl-pose-obj-mod}
For the \textit{Appearance} model in \cref{tb:abl-pose-obj} of the main paper, we use the same pre-trained model with $f_\text{view}$ as the appearance-based gaze direction estimator from the head image for the GFIE dataset \cite{hu2023gfie}. For the GAFA dataset \cite{nonaka2022dynamic}, we use the pre-trained gaze direction module from the GAFA baseline, which also takes the head direction $\mathbf{v}$ from the same model with $f_\text{view}$. Since these appearance-only models lack 3D context input, GA3CE is not applied. For the \textit{Appearance + Pose} model, which does not use object input, a constant latent vector is used as $F_\text{object}$ in the input to the decoder $f_\text{decoder}$ to exclude object information.

\subsection{Depth map processing}  
We utilize the zero-shot metric depth estimator \cite{bhat2023zoedepth} for the RGB-only experiments on the GFIE dataset \cite{hu2023gfie} in \cref{sec:gfie}, as well as for all evaluations on the GAFA dataset \cite{nonaka2022dynamic} in \cref{sec:gafa}.

For backprojecting object positions, as discussed in \cref{sec:contextRepr}, we complete missing regions in the depth maps of the GFIE \cite{hu2023gfie} and the CAD-120 \cite{Koppula2013} datasets using depth completion \cite{liu2022monitored}.

\section{Additional results}
\subsection{Additional results on the GFIE dataset \cite{hu2023gfie}}
\paragraph{Additional qualitative results.}
Additional qualitative results on the GFIE dataset \cite{hu2023gfie} are shown in \cref{fig:supp-gfie-vis}.

\paragraph{Comparison to the baseline GFIE \cite{hu2023gfie} + the SOTA 2D gaze-following method.}
We further compare our proposed approach with the baseline GFIE \cite{hu2023gfie}, replacing its 2D gaze-following module with the latest state-of-the-art method, ViTGaze \cite{song2024vitgaze}. ViTGaze utilizes the large-scale pre-trained model DINOv2 \cite{oquab2023dinov2} as its backbone image encoder. DINOv2 has demonstrated strong capabilities in both object- and scene-level understanding, implicitly learning part-level instance features for diverse categories, including objects and body parts. It also provides a foundation for 3D spatial understanding, such as depth estimation.

It is important to emphasize that the focus of this paper is not to improve or compete with 2D gaze-following methods. As discussed in \cref{sec:intro}, these methods assume different task settings, such as requiring the gaze target to be visible when providing gaze information. In contrast, 3D gaze direction estimation can provide gaze information even when the target is not visible.

We first trained the ViTGaze model and replaced GFIE's 2D gaze-following module during inference. Quantitative evaluation results are shown in \cref{tb:ap-vitgaze}. Incorporating ViTGaze improves 2D gaze-following performance, reducing the L2 distance between ground truth and gaze points by 17\%. Nevertheless, our method still achieves superior results in 3D gaze direction estimation. This is likely because even small errors in the detected gaze point on the image can lead to larger errors in 3D space, especially in the presence of significant depth variation across pixels.

\subsection{Additional results on the CAD-120 dataset \cite{Koppula2013}}
\label{sec:ap-cad120}
\paragraph{Full quantitative results.}
Quantitative results on the CAD-120 dataset \cite{Koppula2013}, including comparisons with other baselines and GFIE \cite{hu2023gfie}, are shown in \cref{tb:cad120_full}. The proposed method outperforms the baseline methods.

\paragraph{Additional qualitative results.}  
Additional qualitative results for the CAD-120 dataset \cite{Koppula2013} are provided in \cref{fig:supp-cad120-vis}.

\subsection{Additional qualitative results on the GAFA dataset \cite{nonaka2022dynamic}}  
Additional qualitative results for the GAFA dataset \cite{nonaka2022dynamic} are shown in \cref{fig:supp-gafa-vis}.

\subsection{Qualitative results for ablation studies}
\paragraph{3D Understanding of Pose and Object.}  
\cref{fig:supp-pose-obj-abl} shows qualitative results for 3D understanding of pose and objects. Blue arrows indicate that combining appearance and pose improves results compared to using appearance alone, marked by magenta arrows. In examples (c) and (d), this combination effectively corrects incorrect estimations, leading to more accurate predictions. Adding object information, represented by green arrows in the \textit{Appearance + Pose + Object} setting, further refines the results, aligning the estimated directions more closely with the ground truth.

\paragraph{Gaze-aware 3D context encoding (GA3CE).}  
\cref{fig:supp-ga3fe-abl} presents qualitative results for GA3CE. Disabling GA3CE (\textit{No GA3CE}) results in less accurate predictions, as indicated by the blue arrows across both datasets \cite{hu2023gfie, nonaka2022dynamic}. Incorporating all proposed components (\textit{ALL}) yields the most accurate results, highlighted by the green arrows.

\subsection{Robustness to view direction}
\label{sec:ap-robustness-v}
To assess robustness to the view direction \( \mathbf{v} \) as a directional prior, we add noise to the ground truth direction so that the resulting \( \mathbf{v} \) has a 3D MAE matching a target value. When the 3D MAE of \( \mathbf{v} \) is 20.0, 30.0, and 40.0 on the GFIE dataset \cite{hu2023gfie}, the model's corresponding 3D MAEs are 11.11, 13.38, and 16.69. On the GAFA dataset \cite{nonaka2022dynamic}, 3D MAEs of 20.0, 29.7, and 38.6 for \( \mathbf{v} \) result in corresponding errors of 18.45, 24.67, and 30.95. The proposed method demonstrates greater robustness on the GFIE dataset \cite{hu2023gfie}, likely due to more frequent close object interactions that aid accurate gaze prediction.

\subsection{Robustness to object absence}
We examine how the model performs when objects are either outside the subject's gaze direction or largely absent from the scene. Objects are removed from the images using the inpainting tool \cite{photoroom_remove_object}, with results shown in \cref{fig:ap-flat-surface}. In (a), the 3D MAE is 14.37 (view direction \(\mathbf{v}\): 15.38), and in (b), the 3D MAE is 10.28 (view direction \(\mathbf{v}\): 27.22). In both cases, the model estimates reasonable gaze directions by leveraging pose and view direction as context cues, even when objects are present but not gazed, or are absent.

\section{Failure cases}
\label{sec:ap-failure}
Typical failure cases are illustrated in \cref{fig:ap-failure-cases}. In (a), the method fails to track gaze shifting from right to left. Since it does not incorporate temporal information, it struggles in situations where pose and object cues alone are insufficient to resolve directional ambiguity. In (b), the model fails to predict a reasonable gaze direction when the view direction \( \mathbf{v} \) contains significant error. As the method relies on \( \mathbf{v} \) as a directional prior in the egocentric frame, it cannot recover from a highly inaccurate view direction.

\section{Limitation and future work}
While our method demonstrates strong performance within the trained domains, as shown in \cref{tb:gfie,tb:gafa}, its generalization to unseen domains leaves room for improvement. This is evident in the results on the CAD-120 dataset \cref{tb:cad120_full}, where Gaze360 \cite{kellnhofer2019gaze360}, despite relying solely on head appearance, outperforms Ours without GFM. Although our method effectively leverages the view direction $\mathbf{v}$ as a prior, it inherits limitations when this prior is suboptimal as discussed in \cref{sec:ap-failure},  particularly in unseen domains. We attribute the suboptimal performance on the CAD-120 dataset to this issue, as seen in the GFIE$_\text{head}$ results in \cref{tb:cad120_full} where the 3D MAE is significantly higher than that of Gaze360 (Gaze360: 20.6, GFIE$_\text{head}$: 27.3). When the view direction is adjusted to match the quality of Gaze360 (view direction 3D MAE: 20.2), following a similar procedure to the experiments in \cref{sec:ap-robustness-v}, our method achieves a 3D MAE of 18.7, outperforming Gaze360. This suggests that our method is more sensitive to the quality of the view direction in unseen domains compared to its robustness in trained domains, as shown in \cref{sec:ap-robustness-v}. As illustrated by the Ours + GFM results in \cref{tb:cad120_full}, incorporating GFM significantly improves performance, even with suboptimal view direction estimates. This highlights a promising direction for improving generalization by integrating gaze-following approaches.

Our method currently assumes the subject's head is visible to the camera without occlusion to enable 3D localization using backprojection with a depth map and head bounding box. Extending this approach to a multi-view setting could address this limitation, and we plan to explore this in future work.  

Additionally, our pipeline does not yet incorporate object semantics, which has been shown to be a promising approach for considering scene context in 2D gaze following \cite{semanticgaze2024}. Expanding our method to include semantic cues alongside object locations is an interesting future direction.  

Finally, the current pipeline focuses on a single subject per scene, consistent with the previous works \cite{hu2023gfie,nonaka2022dynamic}. Future research will explore multi-person scenarios to capture spatial relationships, enabling tasks such as 3D joint attention estimation.

\section{Additional related works}
\paragraph{3D gaze understanding.} Recent works have leveraged 3D gaze information for human motion prediction \cite{zheng2022gimo, lou2024multimodal} and to enhance traffic scene understanding \cite{kong2024wts}. These methods model interactions between gaze and other contextual cues. However, while they assume reliable 3D gaze is given, we instead focus on estimating the 3D gaze itself. Furthermore, the EgoExoLearn dataset \cite{huang2024egoexolearn} has been introduced to bridge asynchronous ego- and exo-centric views for action recognition, incorporating both egocentric and exocentric video alongside gaze data.

\paragraph{Egocentric gaze understanding.} Egocentric video understanding has been extensively studied in the context of action recognition \cite{zhao2023learning, pramanick2023egovlpv2, wang2023ego}. Beyond actions, fine-grained analysis of visual attention in first-person video has been explored through egocentric gaze prediction and gaze-following tasks \cite{li2013learning, zhang2017deep, huang2018predicting, li2021eye, lai2024eye}. Recent studies also leverage egocentric gaze signals as additional prompts for downstream tasks like visual question answering \cite{yan2024voila} and to improve action recognition models \cite{mazzamuto2024gazing}. In contrast to prior work that focuses on understanding the egocentric viewpoint itself, our approach leverages the egocentric view to infer the exocentric gaze of another person.

\paragraph{Ego-exo view understanding.} A number of works have proposed methods to utilize both egocentric and exocentric videos to learn view-invariant semantic representations \cite{li2024egoexo,li2021ego,xue2023learning,park2025bootstrap}. Meanwhile, recent advances in generative models enable direct view transfer between egocentric and exocentric perspectives via image/video synthesis \cite{liu2020exocentric, liu2021cross, luo2024put, xu2025egoexo}. Other approaches incorporate known geometric transformations for cross-view translation \cite{cheng20244diff}. In contrast to these works, our method relies on an explicit geometric view transformation of 3D representations, rather than learning purely view-invariant features or using unconstrained generative view synthesis.

\end{document}